\documentclass{article} % For LaTeX2e
\usepackage{iclr2027_conference,times}
\usepackage{array} 
\iclrfinalcopy

\usepackage{amsmath,amsfonts,bm}

\def\eqref#1{equation~\ref{#1}}
\def\1{\bm{1}}

\DeclareMathAlphabet{\mathsfit}{\encodingdefault}{\sfdefault}{m}{sl}
\SetMathAlphabet{\mathsfit}{bold}{\encodingdefault}{\sfdefault}{bx}{n}

\usepackage{booktabs}
\usepackage[hidelinks]{hyperref}
\usepackage{url}
\usepackage{enumitem}
\usepackage{cleveref}
\usepackage{adjustbox}
\usepackage{subcaption}
\usepackage{multirow}
\usepackage{wrapfig}

\usepackage{xcolor}

\definecolor{RandomColor}{HTML}{4B4C4E}
\definecolor{LinearNoCalColor}{HTML}{4993C0}
\definecolor{LinearCalColor}{HTML}{225188}
\definecolor{BayesianNoCalColor}{HTML}{EF9500}
\definecolor{BayesianCalColor}{HTML}{BA4300}
\definecolor{LLMSameColor}{HTML}{33A02C}
\definecolor{LLMDifferentColor}{HTML}{B2DF8A}

\newcommand{\LinearNoCal}{\textcolor{LinearNoCalColor}{\textsc{Linear (-Calibration)}}}

\newcommand{\BayesianNoCal}{\textcolor{BayesianNoCalColor}{\textsc{Bayesian (-Calibration)}}}
\newcommand{\BayesianCal}{\textcolor{BayesianCalColor}{\textsc{Bayesian (+Calibration)}}}

\title{Rational Use of Time for Social Coordination in Humans and AI}

\author{Xiulin Yang\thanks{Equal contribution.} \\
Georgetown University\\
\texttt{xy236@georgetown.edu} \\
\And
Sirisha Gudavalli$^*$ \\
Johns Hopkins University \\
\texttt{sgudava1@jhu.edu} \\
\And
Jennifer Hu \\
Johns Hopkins University \\
\texttt{jennhu@jhu.edu} \\
}

\usepackage[most]{tcolorbox}
\usepackage{cleveref}
\usepackage{xspace}
\newcommand{\appref}[1]{Appendix~\ref{#1}}
\newcommand{\themind}{\emph{The Mind}\xspace}
\newcommand{\handA}{\ensuremath{\mathbf{A}}\xspace}
\newcommand{\handB}{\ensuremath{\mathbf{B}}\xspace}
\newcommand{\tempo}{\ensuremath{\tau}\xspace}

\newcommand{\change}[1]{{\color{black}#1}}

\newcommand{\todo}[1]{{\color{red}[#1]}}
\definecolor{pos}{HTML}{8B1E2D} 
\definecolor{neg}{HTML}{1F4E79}

\begin{document}

\maketitle
\lhead{Preprint. Under review.}
\begin{abstract}

All social interactions are situated in \emph{time}. Previous studies have shown that humans are able to use implicit temporal cues, such as the duration of a pause, to infer others' mental states. But how do humans use temporal cues for dynamic, fine-grained \change{\textbf{social coordination}}? And can AI systems do the same, without ``experiencing'' continuous real-time as humans do? We study these questions using a cooperative card game task called \themind, which requires players to achieve a shared goal across repeated interactions but forbids explicit communication: the only available signal is the passage of time. 
We evaluate 150 humans, and two classes of AI models which reflect different ways of operationalizing time in artificial systems: computer-use agents (CUAs), and standard text-prompted LLMs. We find that human behavior is best captured by a Bayesian strategy that infers a partner's belief from their timing---and similarly for frontier models. 
However, unlike humans, none of the tested models seem to efficiently adapt their behaviors to their partners' over the course of the game. From a cognitive perspective, our findings suggest that humans perform implicit probabilistic reasoning about continuous-time signals in dynamic social settings, and that such reasoning mechanisms may emerge without real-time experience. Practically,  
our study highlights new challenges for AI systems to coordinate efficiently with other agents by learning to interpret and adapt to their implicit signals in dynamic settings.
% \footnote{Code: \url{https://github.com/hulaboratory/the-mind}.}

% \todo{significance should be more clear: to develop AI assistant in real-life setting...}

% humans actions are best explained by Bayesian belief updating guided by temporal signals, instead of simpler strategies such as mapping card values directly to play times. 

% We find that both humans' and frontier models' However, \todo{overall takeaway should be more clear}in contrast to LLMs and CUAs, humans are better able to adapt their actions to their partners over the course of a game. 
% Our results establish implicit probabilistic reasoning about continuous-time signals as a hallmark of human social cognition \todo{people say it's too strong...}, 

% However, it is unknown how humans rationally use time for complex social coordination. 

\end{abstract}

\section{Introduction}
\label{sec:intro}

Humans draw rich social inferences through all kinds of signals, from linguistic expressions to gestures to the placement of objects.
% . For example, a linguistic expression like ``I'm cold'' might imply that the speaker wants a window to be shut. A gesture toward a door might suggest that it's time to leave. And a backpack placed on a caf\'e table might communicate that someone is planning to return. 
But humans also infer complex information about other agents' mental states
% coordinate to achieve complex cooperative goals in the 
in the \emph{absence} of explicit communicative signal: 
that is, based on \textbf{the passage of time}. For example, the duration of silence before an utterance in a conversation can provide cues into a speaker's trustworthiness or cognitive state \citep[e.g.,][]{christenfeld_does_1995,kohtz_how_2017,ziano_slow_2021}, and solving a complex task in a short amount of time might suggest that the agent had prior knowledge about the task \citep[e.g.,][]{berke_people_2026}. These findings provide evidence that humans use temporal information to draw one-shot social inferences, but we still lack a systematic account of how humans use time as a social signal to dynamically coordinate with other participants. 
% people make inferences about other people's minds based on the amount of time they take to solve a task \citep[e.g.,][]{berke_people_2026}. % The passage of time is not simply the absence of signal---rather, it provides evidence for rich information about people's mental states, enabling efficient inferences across a variety of settings.

It also remains unclear whether AI systems can similarly use temporal information for social coordination. As AI agents are increasingly used in real-time interactions \citep[e.g.,][]{Nakano2021WebGPTBQ,li2026survey}, it seems desirable for AI agents to % Ideally, an AI agent that can cooperate with humans (and other AI agents) should also be able to 
leverage time as an information source about human users or other agents. However, while large language models (LLMs) have been evaluated in diverse social tasks %including collaboration, competition, and coopetition 
\citep{tran2025multi,sidji2024human,zhu-etal-2025-multiagentbench,yim2024evaluating}, most studies have investigated communication based on explicit signals, such as words or non-linguistic symbols. And studies of temporal reasoning typically represent time using explicit timestamps or durations \citep{fatemi2025test, gupta2023temptabqa, wang2024tram}, leaving underexplored how AI systems perform in settings where time unfolds continuously and must be inferred through interaction. % remains underexplored.  %Going beyond these forms of information and incorporating inferences about time could enable more flexible and efficient inferences in a variety of social settings. 

Moreover, investigating time-based social coordination in AI models is not just motivated on practical grounds. It is not even clear whether current AI models are the right \emph{kind of thing} to be able to use time like humans do. LLMs can only iteratively receive discrete prompts, instead of existing in the continuous flow of real-time like biological organisms. Even Computer-Use Agents (CUAs), which can dynamically interact with graphical user interfaces, still perceive the environment through discrete screenshots rather than continuous visual input. It is therefore worth investigating whether such AI systems can perform complex collaborative tasks that require time-based inferences. %---and how such tasks can be reconciled with current LLM architectures.

% And even from the cognitive perspective, most prior work on time-based social coordination in humans has focused on simple tasks, leaving open the question of how---and whether---humans rationally use silence to achieve complex collaborative goals. 

% To address these gaps, 
% In this work, 
Building upon prior work using games to investigate cognition in humans \citep{newell_human_1972,dubey_investigating_2018,allen_using_2024} and AI models \citep{silver_mastering_2016,vinyals_grandmaster_2019,bakhtin_human-level_2022, liang2025llm},
we evaluate humans and models on a cooperative task based on a card game called \themind. In our simplified version of the game, there are two players. On level $n$, each player is randomly dealt $n$ cards from a deck of cards numbered 1 through 100. The players share a cooperative goal: play all dealt cards in ascending order without any communication (no talking, gesturing, winking, etc.). \change{Beyond the cards in their own hands and cards that have been played, the only information players have is the passage of time.} %Suppose, for example, on Level 1 you hold the card 12 and your partner hasn't played their card for an unnervingly long time. This pause might suggest that their card has a relatively high value, increasing your confidence that your 12 is the lowest card. 
This form of coordination lacks any kind of explicit signaling or grounding of symbols into meanings---and yet, people successfully play the game, with minimal training. They even find it fun.\footnote{To quote Paul Dean, the co-founder of board game site and show \emph{Shut Up and Sit Down} \citep{dean_review_2018}: ``Somehow, there are few things in life more satisfying than laying down cards in the right order, something that gets the blood pumping through your veins in a way nothing else quite can, something that gives you the feeling of being a true hero, a feeling that you know few others will ever really experience in their lives. \themind is one of the purest, simplest and best games I have ever played.''}

% We use \themind to investigate the question: \textbf{Whether and how can systems that do not experience continuous real time use the passage of time, or silence, as social information during repeated interactions, as humans do?} Specifically, we aim to answer the following subquestions:
% \begin{itemize}[itemsep=0pt, topsep=0pt, parsep=0pt]
% \item \textbf{Performance:} How do AI models perform compared with humans?
%  \item \textbf{Strategy:} Do AI models and humans share the same underlying reasoning strategy?
% \item \textbf{Convergence:} Can AI models and humans adjust their behavior based on their partner's behavior, and do they converge on a shared convention over repeated interactions? 
% \item \textbf{Behavior Consistency:} Do AI models show consistent behavior when being prompted with different time granularities? 
% \end{itemize}

Using \themind, we \change{focus on one underexplored aspect of social interaction: coordination in the absence of explicit signals. We} investigate two main questions: How do humans use temporal cues for dynamic social coordination? And can AI systems also coordinate using such cues, despite not ``experiencing'' continuous real time? We break these questions down into the following subquestions:
\begin{itemize}%[itemsep=0pt, topsep=0pt, parsep=0pt]
% \item \textbf{RQ1 (Performance):} How do AI models perform compared with humans?
% \item \textbf{RQ2 (Strategy):} Do AI models and humans use the same underlying reasoning strategy?
% \item \textbf{RQ3 (Adaptation):} Do AI models and humans adjust their behavior based on their partner's behavior, and do they converge on a shared convention over repeated interactions? 
% \item \textbf{RQ4 (Behavior Consistency):} Do AI models show consistent behavior when being prompted with different time granularities? 
\item \textbf{RQ1 (Performance):} How successfully and efficiently do humans and models perform? 
\item \textbf{RQ2 (Strategy):} What cognitive strategy best characterizes human and model behavior?
\item \textbf{RQ3 (Adaptation):} Do humans and models adapt their behaviors to increase success and efficiency with their partner across repeated interactions?
\item \textbf{RQ4 (Behavior Consistency):} Do models show consistent behavior when prompted to reason about time at different granularities? 
\end{itemize}

% \begin{enumerate}
%     \item[\textbf{RQ 1:}] \textbf{How do people coordinate using time?} We hypothesize that people use time/silence as a continuous stream of evidence that enables Bayesian inference about other players' cards, while simultaneously establishing conventions without any explicit communication.
%     \item[\textbf{RQ 2:}] \textbf{Can AI models reason about time to achieve complex cooperative goals like humans do?} Standard LLMs operate through discrete, pausable, token-based ``steps'', in contrast to the flow of real-time that supports biological experience. Can this mode of existence be used to approximate something like real-time cooperation? 
% \end{enumerate}

% \todo{First, do people form signaling conventions based on the passage of time? Time-based conventions already exist in everyday life. For example, people have shared knowledge of how long to wait to hold the door for someone behind us; how long certain pauses should last before it verges into awkward territory; and so on. But the kind of conventions formed during \themind are more precise, and are formed across a much shorter time span.

% Second, how do people use incremental evidence? \todo{elaborate on linear vs Bayesian strategies} We hypothesize that people use time/silence as a continuous stream of evidence that enables Bayesian inference about other players' cards, while simultaneously establishing conventions without any explicit communication.}

We implement five types of cognitive strategies as fixed partners, and evaluate how three types of players interact with them: 
% humans, Computer-Use Agents (CUAs), and LLMs. 
% To investigate these questions, we evaluated three types of players: 
(1) humans; (2) CUAs, which operate through a real-time graphical user interface; and (3) standard LLMs, which can only reason about time through prompts.
% Through these three experiments, we aim to (i) 
% %analyze which partners lead humans to be most accurate and efficient; 
% understand how humans use temporal information for complex cooperation;
% (ii) investigate whether agents that operate through a real-time graphical user interface can still make efficient use of temporal signals for coordination; and (iii) test whether models that reason about time explicitly through prompts %(e.g., choosing whether to wait for 1 second) 
% % can perform the task as effectively as humans do by directly experiencing time.
% can effectively perform the task.
We find that both humans' and frontier LLMs' behaviors are best characterized by a Bayesian strategy that reasons about time to infer the partner's state (\Cref{sec:strategy}). However, although frontier LLMs can match or even outperform humans in task success (\Cref{sec:overall}), humans show greater adaptation to their partners' behavior over the course of the game (\Cref{sec:convention}). LLMs' performance is also sensitive to how temporal information is represented in the prompt (\Cref{sec:time}). 

Our results suggest that humans perform implicit probabilistic reasoning about continuous-time signals in dynamic social \change{coordination} settings. Frontier LLMs may have internalized the same rational inferential strategy that humans use to coordinate under temporal uncertainty. However, this alignment does not appear to be driven by the same experience-driven, online adaptation to a specific partner that characterizes human coordination. LLMs may already outperform humans in aggregate coordination outcomes, but efficiently and consistently adapting to a partner during an interaction remains an open challenge---especially when using implicit temporal signals without linguistic grounding.
% one that current scaling trends alleviate but do not fully resolve.

% \todo{summarize results and connect to bigger picture -- highlighting time perception and time-based inferences as an open challenge for AI, potentially requiring new architectures}

% Our contributions are as follows:
% \begin{enumerate}
%     \item Novel dataset of human behaviors in a cooperative task involving communication with silence
%     \item Novel 
% \end{enumerate}

\begin{figure}[t]
    \centering
    \includegraphics[width=0.9\linewidth]{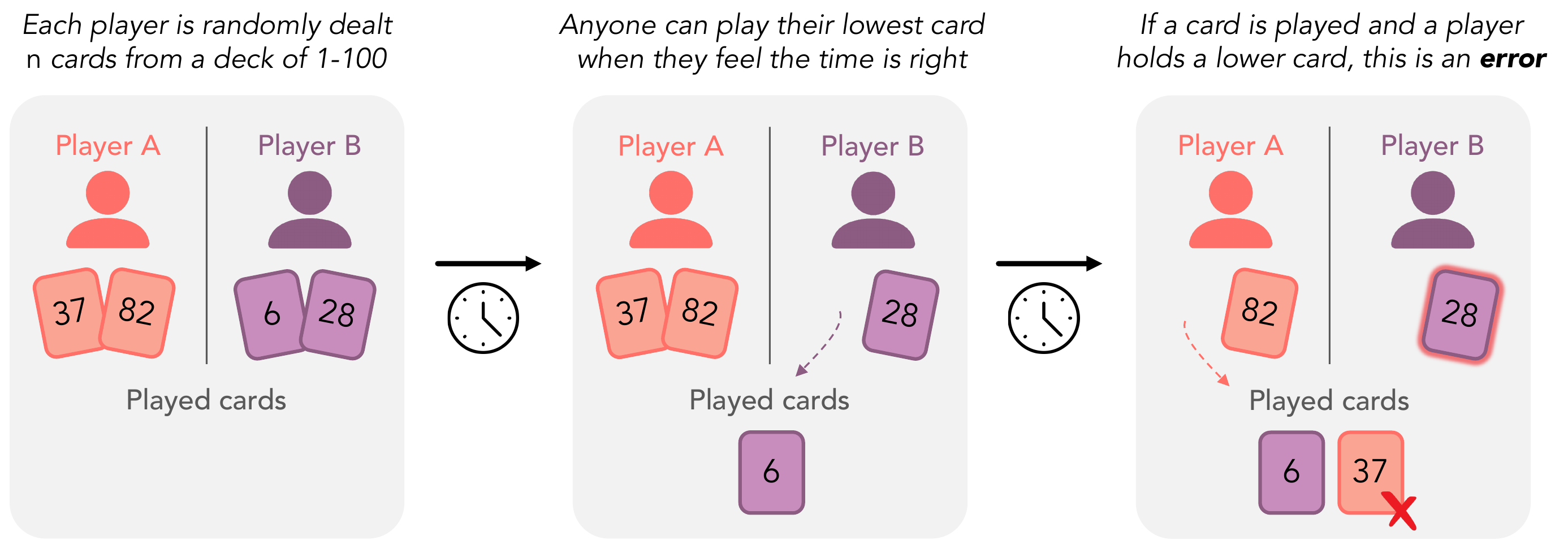}
    \caption{Overview of task requiring time-based inference (based on \themind). Each player holds their own private cards (dealt uniformly at random from 1-100), and can see the previously played pile of cards. The players jointly share the goal of playing all dealt cards in ascending order, from lowest to highest, without any communication. Players simply play when they feel the time is right.} %There are no turns. Players play their lowest card when they feel the time is right.}
    \label{fig:overview}
\end{figure}

\section{Related Work}
\label{sec:rw}
Humans can rapidly establish shared conventions with a stranger, even when they have no prior coordination history \citep[e.g.,][]{clark1986referring,garrod1994conversation}. %Through interaction, interlocutors can develop shared knowledge that facilitates coordination on a new task. 
Importantly, convention formation does not rely exclusively on linguistic communication. Humans can coordinate through discrete linguistic cues \citep{steels2005coordinating}, discrete non-linguistic cues \citep{misyak2016instantaneous}, continuous non-linguistic cues \citep{chen2025discrete}, and body gestures \citep{pezzulo2019body}. 
%The ability to rapidly adapt to new interaction partners and tasks is also desirable for AI systems. 
A growing body of work has also investigated convention formation in artificial agents from different perspectives, including coordination games \citep{sidji2024human,shih2021critical} and pragmatic reasoning \citep{yu-etal-2026-pragmatic}. Various benchmarks have also evaluated current agent systems' collaboration ability in complex tasks \citep{zhu-etal-2025-multiagentbench,shih2021critical,yim2024evaluating,yee2026benchmarking}. Other studies have specifically examined LLM collaboration in games under implicit or communication-constrained settings \citep{liang2025llm,buscemi2026numbers}.
% . More recent studies start to evaluate LLMs under limited communication situation using games \citep{liang2025llm}. 

%\paragraph{Reasoning on Time in Human \& LLM studies} 
% \paragraph{Temporal reasoning}
% Reasoning purely from the passage of time has received comparatively little attention in the human cognition literature. 
Another relevant line of work has investigated temporal inference in humans and LLMs. 
%Prior studies have explored how 
Humans infer other people's beliefs or preferences from response or decision time \citep{gates2021rational,bavard2024humans,konovalov2023decision}. Recently, \citet{berke_people_2026} showed that people can infer how %much of a pause 
long another agent spent distracted %(as opposed to actively thinking) 
purely from the duration of a pause, and that this inference is well captured by a Bayesian model of computation. 
Temporal reasoning has also received increasing attention in LLMs, with a growing number of benchmarks designed to assess models’ ability to reason about temporal information \citep[e.g.,][]{fatemi2025test,islakoglu2025chronosense,gupta2023temptabqa,wang2024tram,chu2024timebench,fatemi2025test,islakoglu-kalo-2025-chronosense,wei2026time}. More recently, \citet{sehgal2026real} show that LLMs exhibit fragile adaptation to wall-clock time pressure in negotiating settings, while \citet{wang-etal-2025-discrete} find that LLMs can associate token counts with the passage of continuous time.  However, these evaluations primarily evaluate whether LLMs can reason about explicitly specified and static temporal information, leaving how LLMs represent and track the passage of time during dynamic interaction underexplored. 
% Temporal reasoning has also received increasing attention in LLMs, with a growing number of benchmarks designed to assess models’ ability to reason about temporal information \citep[e.g.,][]{fatemi2025test,islakoglu2025chronosense}. Many benchmarks represent time using discrete timestamps and evaluate temporal reasoning based on information available from world knowledge or textual contexts \citep[e.g.,][]{gupta2023temptabqa,wang2024tram,chu2024timebench,fatemi2025test,islakoglu-kalo-2025-chronosense}. 
% % For example, TempTabQA \citet{gupta2023temptabqa} evaluates temporal understanding and reasoning over semi-structured tables derived from Wikipedia. Similarly, \citet{wang2024tram,chu2024timebench,fatemi2025test,islakoglu-kalo-2025-chronosense} introduce benchmarks that vary in their data sources, domains, and question formats. 
% However, these benchmarks primarily evaluate whether LLMs can reason about explicitly specified temporal information, leaving how LLMs represent and track the passage of time during ongoing interaction underexplored. 

\section{Methods}
\label{sec:method}

\subsection{Overview of Experiments}

Our task is based on a cooperative card game called \themind (see \Cref{fig:overview}). There are two players in the game, and a deck of cards numbered 1 through 100. The game proceeds across ten levels. On level 1, each player is randomly dealt 1 card from the deck; on level 2, each player is randomly dealt 2 cards from the new deck; and so on. On each level, the players share a cooperative goal: play all dealt cards in ascending order (lowest to highest). \change{During the game, players know their own cards, the cards that have already been played (and hence how many cards their partner still holds), and the time elapsed since the last play, but they never see their partner's cards. Accordingly, in our game, by the partner's \textit{mental state}, we mean their private knowledge of their own cards, i.e., the value of their lowest card.}  Critically, there are no turns, and players cannot communicate with each other. Players simply play a card when they feel that it is the next lowest. If a player plays a card that is higher than the lowest card held by another player, this counts as an error, and all the lower card(s) are automatically played.  
% The only information players have is the passage of time. 

% As shown in \Cref{fig:overview}, the goal of the players is to play out the card from an ascending order. However, during the game, the players cannot communicate with each other. 

% Our experiments are designed as follows. 
We evaluate three types of players on the task: humans, CUAs, and LLMs. In each experiment, we pair the agent of interest against artificial partners that implement different kinds of cognitive strategies. By manipulating the partner strategies, we can perform controlled analyses on what kind of partners leads to the best cooperation with each agent. \change{Note that in our experiments, we did not pair humans with humans for two main reasons. First, identifying a player’s strategy and measuring adaptation requires the partner’s strategy to be known. When both players’ strategies are unknown and co-evolving, it becomes difficult to identify which player is adapting to whom, and the strategy-fitting analysis becomes much noisier. Second, a shared set of partners makes humans, LLMs and CUAs comparable with each other in their adaption and fitting strategies. 
}

% and test whether each agent changes their own strategy in response to different partners.
We describe the rule-based partner strategies in more detail in \Cref{sec:rule-based}, and then describe the methods of each experiment in \Cref{sec:expts}.

\subsection{Rule-based Partners} \label{sec:rule-based}

% We pair each agent (humans, CUAs, and LLMs) with five kinds of partners that implement rule-based strategies. 
% There are three main types of cognitive strategies: Linear, Bayesian, and Random. The Linear and Bayesian player types can additionally either perform calibration (i.e., adjust their speed based on their partner's hesitation time) or not perform calibration. 
We pair each agent (humans, CUAs, and LLMs) with five kinds of partners implementing different cognitive strategies: Linear +/-Calibration, Bayesian +/-Calibration, and Random. These strategies differ along two separate dimensions: first, the way that time is used (Linear, Bayesian, or Random), and second, whether play speed is adjusted based on their partner's hesitation time (+/-Calibration). The Linear strategy is motivated by the evidence that adults represent numbers linearly \citep{siegler2003development}, making a linear mapping between the card value and waiting time a plausible simple default. The Bayesian strategy is motivated by prior work that show human temporal reasoning or perception are well described by Bayesian models \citep[e.g.,][]{berke_people_2026,jazayeri2010temporal}. The Random strategy is taken as a baseline.
Additional details are provided in \appref{app:players}. 

At the start of level $n$, player A is dealt a hand of $n$ cards $\langle a_1, a_2, \cdots, a_n\rangle$, where $a_1 < a_2 < \dots < a_n$, and player B
similarly is dealt $\langle b_1, b_2, \cdots, b_n\rangle$. Cards are played one at a time. 
Let $c_i$ denote the $i$-th card played, and use $t$ to denote elapsed wall-clock time since the most recent card was played. 
% \sg{Let $c_i$ denote the $i$-th card played (in play order), and let $j$ denote the number of cards played so far.} 
We write $c^{(t)}$ to denote the most recently played card at time $t$,
and $\mathbf{c}^{(t)} = \{c_1, c_2, \dots, c^{(t)}\}$ to denote the history of cards played until time $t$.
% At the beginning of a round and before the first card is played, we set
% the card value $0$ (there is no $0$ card in the deck);
% $\mathbf{c}^{(j)} = \{c_1, c_2, \ldots, c^{(j)}\}$ denotes the history of
% played cards, visible to both players; $\mathbf{A}^{(j)} \subseteq \mathbf{A}$
% denotes player A's remaining cards (known only to A); and
% $\mathbf{B}^{(j)} \subseteq \mathbf{B}$ denotes player B's remaining cards
% (known only to B).  
We also write $a_1^{(t)}$ to denote player A's lowest remaining card, and $b_1^{(t)}$ to denote player B's lowest remaining card at time $t$. The $t$ ``clock'' is reset to 0 at the start of each round and each time a card is played.
% and by $b_1^{(j)}$ player B's lowest remaining card. 

% We distinguish this play-order indexing from wall-clock time. \sg{We distinguish the play-order index $j$ from elapsed time $t$.} Throughout,
% $t \geq 0$ denotes the time (in seconds) elapsed since the most recent card
% was played; this clock is reset to zero at the start of each round and each
% time a card is played.  Each decision rule below specifies the elapsed time at which a player plays
% their lowest remaining card; we denote this play time by $t^*$. 
% When we need the timing of
% a particular past play, we write $t_i$ for the duration of the silence that
% preceded the $i$-th card, i.e., the value of the clock at the moment $c_i$ was
% played.

\subsubsection{Strategy types} \label{sec:strategies}
\paragraph{Linear player.}

The linear player waits a fixed amount of time proportional to the gap between the most recently played card $c^{(t)}$ and their lowest card. In other words, they use a fixed scaling factor---the ``tempo'' \tempo---to determine when to play their lowest card. For a given tempo \tempo, player A waits to play their lowest card $a_1^{(t)}$ at elapsed time $\tempo \cdot \left(a^{(t)}_1 - c^{(t)}\right)$ (and similarly for player B). % Player A's decision rule is therefore to play $a_1^{(t)}$ at the elapsed time:
% \begin{equation}
%     t^* = \tau \cdot \left( a_1^{(t)} - c^{(t)} \right)
%     \label{eq:counting-decision-rule}
% \end{equation}
% and player B symmetrically plays $b_1^{(t)}$ at $t^* = \tau \cdot \left( b_1^{(t)} - c^{(t)} \right)$.

If both players share the same tempo \tempo, and follow this strategy perfectly and deterministically, then this strategy is guaranteed to succeed. However, this strategy is sensitive to miscalibration of \tempo, potentially leading to mistakes. Separately, this strategy ignores the \emph{information} in the other player's silence, potentially leading to inefficiency (i.e., wait times that are longer than necessary). % This issue is alleviated by the Bayesian strategy.

\paragraph{Bayesian player.}

Instead of using a fixed mapping between card values and time units, the Bayesian player rationally uses their partner's silence as \emph{evidence}. % incrementally update their belief about whether their partner's smallest card is larger than their own smallest card. 
Without loss of generality, we focus on player A's perspective. 
% Let $\mathbf{C^{(t)}}$ be a random variable denoting the history of cards played until time $t$, and 
% , with possible values 
Let $B^{(t)}_1$ be a random variable denoting player B's lowest card at time $t$, and let $T_B$ be a random variable denoting the time at which player B plays their lowest card. This means that at time $t$, player A has observed that $T_B > t$. Player A can use Bayesian inference to compute the probability that player B's current lowest card $B^{(t)}_1$ is equal to some value $k$:
{\begin{align}
    % P\left(B^{(t)}_1 = k \mid T_B > t, \mathbf{C^{(t)}} = \mathbf{c}^{(t)}\right) &= \frac{P\left(T_B > t \mid B^{(t)}_1 = k, \mathbf{C^{(t)}} = \mathbf{c}^{(t)}\right) \cdot P\left(B^{(t)}_1 = k \mid \mathbf{C^{(t)}} = \mathbf{c}^{(t)}\right)}{P\left(T_B > t \mid \mathbf{C^{(t)}} = \mathbf{c}^{(t)}\right)}
    % P\left(B^{(t)}_1 = k \mid T_B > t, \mathbf{C^{(t)}} = \mathbf{c}^{(t)}\right) &\propto P\left(T_B > t \mid B^{(t)}_1 = k, \mathbf{C^{(t)}} = \mathbf{c}^{(t)}\right) \cdot P\left(B^{(t)}_1 = k \mid \mathbf{C^{(t)}} = \mathbf{c}^{(t)}\right)
    P\left(B^{(t)}_1 = k \mid T_B > t, \mathbf{c}^{(t)}\right) &\propto P\left(T_B > t \mid B^{(t)}_1 = k, \mathbf{c}^{(t)}\right) \cdot P\left(B^{(t)}_1 = k \mid \mathbf{c}^{(t)}\right)
\end{align}}%
To compute the prior probability that player B's lowest card $B^{(t)}_1$ is equal to some card value $k$, player A assumes a uniform draw from all cards that are not in their hand and have not been played.
% can use the history $\mathbf{c^{(t)}}$, as well as the cards they hold, to determine the possible cards that player B can hold. The players assume a uniform draw from the remaining possible cards.
To compute the likelihood of player B waiting at least until time $t$ to play card with value $k$, we model each player's play time as a random variable drawn from a hazard process that depends on their card value. That is, $\left(T_B \mid B = k, c^{(t)} \right) \sim\text{Weibull}\left(\gamma, \frac{1}{\tempo \cdot (k - c^{(t)})}\right)$, where \tempo is the tempo parameter and $\gamma$ is a parameter controlling the change of the hazard rate. If $\gamma > 1$, the hazard rate increases
over time, which captures impatience in the game. A full derivation is provided in \appref{app:bayesian}.

% \paragraph{Decision rule}

% Player A waits to play until they are confident enough that they hold the lower card. The longer the silence, the more that confidence grows. So player A plays at the following time:
% \begin{align}
%     t^* = \inf \{t \geq 0 : P(B^{(t)}_1 > a^{(t)}_1 \mid T_B > t) \geq \theta\}
% \end{align}
% where $\theta > 0$ is a decision threshold. The lower $\theta$ is, the less confidence player A needs before they decide to play.

Player A waits to play until their belief that they hold the lower card exceeds a confidence threshold $\theta > 0$. In other words, they play at the smallest $t$ such that $P(B^{(t)}_1 > a^{(t)}_1 \mid T_B > t) \geq \theta$.
% So player A plays at the following time:
% \begin{align}
%     t^* = \inf \{t \geq 0 : P(B^{(t)}_1 > a^{(t)}_1 \mid T_B > t) \geq \theta\}
% \end{align}

\paragraph{Random player.}

As a baseline, we consider a random player who decides whether to play or wait every second with probability 0.5. 

\subsubsection{Tempo calibration} \label{sec:tempo-calibration}

% Separate from the use of evidence is how tempo is calibrated between players. The core intuition is that the game requires players to somehow form a ``tempo'' convention, without explicitly communicating. We will denote this tempo with the variable $\tempo > 0$: a smaller $\tempo$ means that everyone plays more quickly; a larger $\tempo$ means that everyone plays more slowly. 

% \todo{need to improve this} The Linear, Bayesian, and Random strategies specify how a player \emph{uses} time. On a separate dimension, a strategy for playing the game also involves \emph{calibration between players}. Intuitively, the game requires players to form a ``tempo'' convention without explicitly communicating. We will denote this tempo with the variable $\tempo > 0$.
The Linear and Bayesian players both make use of a ``tempo'' parameter $\tempo > 0$. Intuitively, the tempo of the game serves as a scaling factor between card value units and time units. A smaller $\tempo$ means that everyone plays more quickly; a larger $\tempo$ means that everyone plays more slowly. 
A \textbf{-Calibration} player begins with a fixed tempo $\tempo_0$ and does not update it throughout the course of the game. A \textbf{+Calibration} player begins with some $\tempo_0$ and updates it after each played card.
% We can define a single-observation MLE in the following way.  
After each card $c_i$ is played at time $t_i$ since the previous play, the players observe a new data point and can immediately update their estimate of $\tau$. The single-observation MLE implied by this data point is $\hat\tau_i = \frac{t_i}{c_i - c_{i-1}}$, where $c_{i-1}$ is the previously played card.
% The single-observation MLE implied by this
% data point is:
% \begin{equation}
%     \hat\tau_i = \frac{t_i}{c_i - c_{i-1}}
%     \label{eq:mle-tempo}
% \end{equation}
% where $c_{i-1}$ is the previously played card.
For players with tempo calibration, we define the players as taking an exponential moving average, weighted toward recent observations: $\hat{\tempo}_i = (1-\eta)\cdot \hat{\tempo}_{i-1} + \eta \cdot \frac{t_i}{c_i - c_{i-1}}$,
% \begin{align}
%     \hat{\tempo}_i = (1-\eta)\cdot \hat{\tempo}_{i-1} + \eta \cdot \frac{t_i}{c_i - c_{i-1}}
% \end{align}
where $\eta \in (0,1)$ controls the rate of adaptation to recent observations.

\subsection{Experimental Methods} \label{sec:expts}

\subsubsection{Experiment 1: Humans vs.~Rule-based partners} \label{sec:expt-human}
% We randomly generated 10 game seeds, each defining 10 levels with 2 pairs of hands per level. Each seed was evaluated under 5 different conditions: Bayesian players with calibration enabled or disabled, Linear players with calibration enabled or disabled, and a random player as a baseline, resulting in 50 seed-by-condition combinations. For all Bayesian and Linear players, we set \tempo to 0.5, $\gamma$ to 2, and $\theta$ to 0.85. For players with calibration enabled, we set $\eta$ to 0.3. These parameter values were selected based on the results and feedback from the pilot study.

% For each seed-by-partner-type combination, we recruited 3 human participants through Prolific, resulting in 150 participants in total. Participants played the game with rule-based players through a web-based game interface. Before playing, participants provided informed consent and received detailed instructions on the game rules and the layout of the game interface (see \Cref{app:game_layout}). They were then required to answer three comprehension questions based on the instructions before proceeding to the game demo. To encourage efficient cooperation, we provided a bonus up to \$2 based on the game during and success rate apart from the base payment. More details can be found in Appendix~\ref{human-exp} 

We randomly generated 10 game seeds, each defining 10 levels with one pair of hands per level. Each seed was evaluated under 5 different partner conditions: Bayesian +/-Calibration, Linear +/-Calibration, and Random, resulting in 50 seed-by-partner combinations. For all Bayesian and Linear players, we set \tempo to 0.5, $\gamma$ to 2, and $\theta$ to 0.85. For players with calibration enabled, we set $\eta$ to 0.3. These parameter values were selected based on the results and feedback from the pilot study.

For each seed-by-partner combination, we recruited 3 human participants through Prolific, resulting in 150 total participants. Participants played the game with the rule-based partners through a web-based game interface. Before playing, participants provided informed consent and received detailed instructions on the game rules and the layout of the game interface. They were then required to answer three comprehension questions based on the instructions before proceeding to the game. To encourage efficient cooperation, we provided a bonus up to \$2 based on the game duration and success rate apart from the base payment. Experiment details can be found in \appref{human-exp}.

\subsubsection{Experiment 2: Computer-use agents (CUAs) vs.~Rule-based partners} \label{sec:expt-cua}

% In this experiment, we evaluate three frontier CUAs: \texttt{Gemini-3.7-flash}, \texttt{GPT-5.6-luna}, and \texttt{GPT-6-astra}. We focus on CUAs because they are specifically designed to perform real-time, interactive tasks in graphical user interfaces. This makes them a natural choice for investigating whether AI agents can make use of temporal information during social coordination. Moreover, because both CUAs and human participants interact with the same game interface, their behavior can be compared under more closely matched interaction environments.

% Throughout the paper, we refer to the agents as \textit{[model name]-CUA} to distinguish them from their underlying language models. All three CUAs interact with the game under the same experimental conditions and through the same interface used by human participants. However, due to the computational cost of running CUAs, we run each CUA only once for each game seed-partner-type combination, resulting in 50 data points per CUA. Before playing, the CUAs read the game instructions and complete comprehension questions to ensure that they understand the task. We also provide the game rules in the system prompt throughout the interaction.

CUAs are built on large vision-language models. They take multimodal content as input and can use computer tools to interact with and operate GUIs in real-time, making them a natural choice for investigating whether AI agents can use temporal information during social coordination. %Moreover, because both CUAs and human participants interact with the same game interface, their behavior can be compared under nearly identical settings. All three 
CUAs interact with the game under the same experimental conditions and through the same interface used by human participants, including instructions and comprehension questions. Therefore, their interaction time is measured using wall-clock time, as it is for human participants.

We evaluate three frontier CUAs: \texttt{Gemini-3.7-flash} \citep{team2023gemini}, \texttt{GPT-5.6-luna}, and \texttt{GPT-6-astra}.\footnote{We refer to the CUA models as \textit{[model]-CUA} to distinguish them from their underlying language models.} Due to computational costs, we run each CUA only once for each game seed-partner combination, resulting in 50 (10 seeds$\times$5 partners) data points per CUA. 

\subsubsection{Experiment 3: LLMs vs.~Rule-based partners} \label{sec:expt-llm}

Finally, we consider standard LLMs which can only ``use'' time by reasoning about it in text-based prompts. We evaluate 10 instruct-tuned LLMs from 5 families, spanning across sizes and open-/closed-source settings: \texttt{GPT-6-astra} and \texttt{GPT-5.6-luna}; \texttt{Gemini-3.7-flash}; \texttt{GPT-oss-120b} and \texttt{GPT-oss-20b} \citep{openai2025gptoss120bgptoss20bmodel}; \texttt{Qwen3-32B} and \texttt{Qwen3-8B} \citep{qwen3technicalreport}; and \texttt{LLaMa-3.1-70b}, \texttt{LLaMa-3.1-8b}, and \texttt{LLaMa-3.1-1b} \citep{Touvron2023LLaMAOA}. The experiment uses the same conditions as the CUA experiment, with 50 data points per model. We also pair LLMs with the same LLM model to see how well they cooperate with each other. For LLMs that show performance comparable to humans, we additionally pair them with \texttt{Gemini-3.7-flash} to examine how they coordinate with a higher- or similarly performing partner. Conversely, we pair \texttt{Gemini-3.7-flash} with \texttt{gpt-oss-120b} to examine how it coordinates with a lower-performing partner. Hyperparameter details can be found in \Cref{tab:hyperparameters}.

As LLMs cannot directly perceive the passage of time, % and have unrealistically large working memory compared with humans \citep{wang2026simulating}, 
we prompt the LLM at simulated time steps with the current game state and ask the model whether it decides to wait (\texttt{WAIT}) or play the lowest card (\texttt{PLAY}). %To simulate humans' limited working memory, which is often characterized as being able to maintain approximately 7 $\pm$ 2 chunks of information \citep{miller1956magical}, e restrict the game history provided to the LLM to the five most recent card plays as a simple approximation of this memory limitation. 
% To test the robustness of LLMs to different prompts and whether LLMs' behavior will change if prompted at different time-granularity, we prompt the LLMs under three main prompt conditions. First, in the \textbf{time-based} condition, the elapsed time after each wait is explicitly specified. We consider three temporal granularities: in each prompt, the elapsed waiting time is updated in increments of 1 second, 0.5 seconds, or 0.25 seconds. We additionally include a fourth condition in which the model is simply asked to generate an estimated waiting time as an absolute value based on the current game state. Second, in the \textbf{wait-based (count)} condition, the number of times the model has waited is explicitly provided as a numerical value. Third, in the \textbf{wait-based} condition, only the history of previous decisions is presented, without an explicit numerical count (e.g., \textit{\texttt{WAIT, WAIT, WAIT}} if the LLM has decided to wait three times). Overall, when the decision process is not explicitly grounded in time, these conditions allow us to investigate what temporal representation LLMs implicitly encode, and whether this representation is consistent when explicit numerical information is present versus absent. The latter two conditions allow us to examine whether LLMs rely on explicit numerical information when reasoning about elapsed time. We test these prompts with all LLMs except \texttt{GPT-6-astra} (only the prompt with one second) due to computational costs.
To assess whether models behave consistently with different temporal granularities, we consider the following conditions: (i) \textbf{Time-based}: The elapsed waiting time is explicitly provided. We use four variants: the time is updated in increments of 1 second, 0.5 seconds, or 0.25 seconds, or the model directly generates an absolute waiting-time estimate based on the current game state. (ii) \textbf{Wait-based (count)}: The number of previous \texttt{WAIT} decisions is provided as a numerical value. (iii) \textbf{Wait-based (action)}: The history of previous decisions is provided as a sequence of tokens (e.g., \texttt{WAIT, WAIT, WAIT}). The Wait-based conditions test whether LLMs can perform the task by relying on numerical information, even when it is divorced from real time units.\footnote{In order to ``play'' with LLMs in the Wait-based conditions, the rule-based partners estimate the elapsed waiting time from the number of \texttt{WAIT}s, assuming that each \texttt{WAIT} token corresponds to one second.} We evaluate all LLMs under these conditions except \texttt{GPT-6-astra}, for which we use only the 1-second Time-based condition due to costs. Details of prompts can be found in \Cref{app:prompt} and \Cref{tab:prompt-conditions}.  

% Across all conditions, only the prompt is varied, while the partner’s behavior remains unchanged. 

\section{Results}
\change{We now address the four research questions laid out in the Introduction. Note that, unlike humans and CUAs, text-based LLMs do not experience wall-clock time and are prompted at simulated time steps. Comparisons of absolute timing between LLMs and the other players in RQ1 should therefore be interpreted with caution. The remaining research questions are not affected by this difference, as they rely on comparisons \emph{within} each setting: strategy fitting is performed on each game record (RQ2), adaptation is measured as change over the course of a game (RQ3), and the effect of temporal representation is compared across prompts within a single model (RQ4).}
\subsection{RQ1: Overall Performance of Humans, CUAs, and LLMs}
\label{sec:overall}

\begin{figure}
    \centering
    \includegraphics[width=\linewidth]{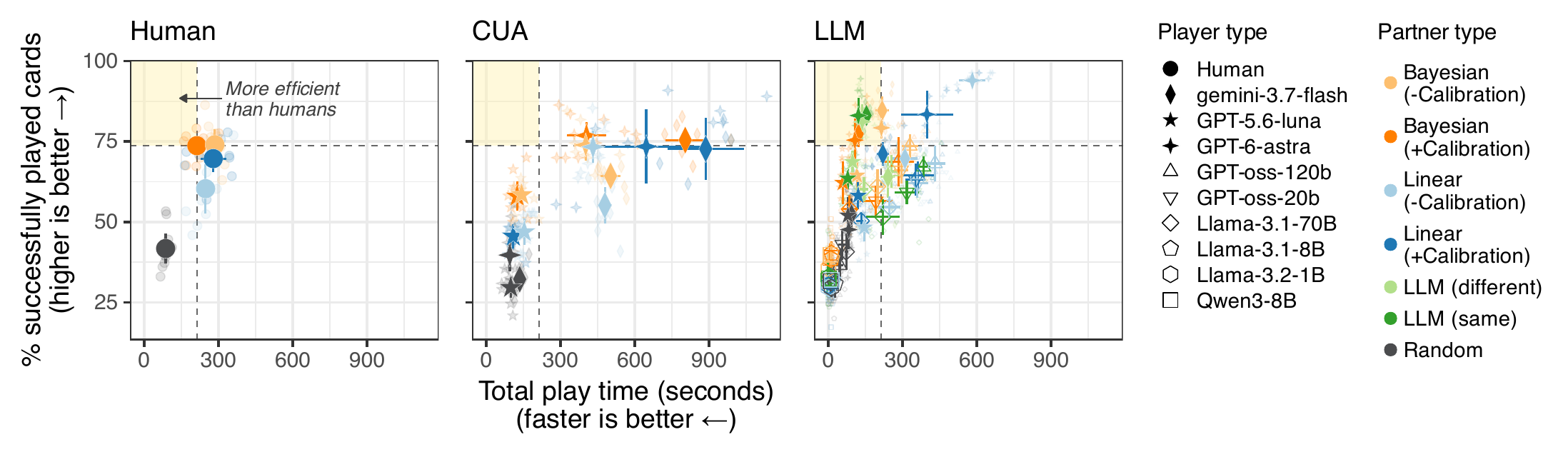}
    \caption{Success rate versus total play time, for humans, CUAs, and LLMs with error bars showing 95\% CI. Total play time is directly measured in wall-clock time for humans and CUAs, and indirectly computed based on prompt responses from LLMs. Opaque large markers represent means across game seeds; more transparent smaller markers represent individual game seeds.}
    \label{fig:overall_performance}
\end{figure}
% \begin{figure}[t]
%     \centering
%     \begin{subfigure}[t]{\linewidth}
%         \centering
%         \includegraphics[width=\linewidth]{figs/human_llm_use_combined.pdf}
%         % \caption{Overall performance across humans, LLMs, and CUAs.}
%         \caption{Success rate versus total play time, for humans, CUAs, and LLMs. Total play time is directly measured in wall-clock time for humans and CUAs, and indirectly computed based on prompt responses from LLMs. \todo{explain in caption what transparency and size mean. also, error bars.}}
%         \label{fig:overall_performance}
%     \end{subfigure}
%     \vfill
%     \begin{subfigure}[t]{\linewidth}
%         \centering
%         \includegraphics[width=\linewidth]{figs/top10_efficiency.pdf}
%         \caption{Top-20 most efficient model-partner combinations.}
%         \label{fig:efficiency}
%     \end{subfigure}

%     % \caption{Overall performance and efficiency comparison across player types.}
%     \caption{Task performance across experiments. Filled shapes = humans or closed-source models.}
%     \label{fig:performance-time}
% \end{figure}

We first analyze overall performance, using two complementary measures: \textbf{game duration}, defined as the time elapsed between the start and end of the game, and \textbf{success rate}, defined as the number of successfully played cards divided by the total number of cards (110). An ideal player would successfully play as many cards as possible while completing the game in as little time as possible.

The results are presented in \Cref{fig:overall_performance}.\footnote{For LLMs, we report results from the time-based (1s) prompt. We compare across prompts in \Cref{sec:time}.} We observe three noteworthy findings. First, across all player types---humans, CUAs, and LLMs---the best performance (i.e., trade-off between success rate and game duration) is generally achieved with Bayesian partners. This may suggest that people's (and models') default expectations about partner behavior are more consistent with a Bayesian strategy than alternatives like the Linear strategy. Second, frontier models can achieve performance comparable to, and in some cases exceeding, that of humans. The rightmost subplot shows that certain LLM players achieve both higher success and faster play times than the most efficient human games.\footnote{To jointly evaluate success rate and game duration, we compute an ``efficiency'' score by z-scoring time and success and computing $-0.5 \times time_z + 0.5 \times success_z$. We visualize top-20 most efficient player-partner combinations in \Cref{fig:top20}. Closed-source LLMs are among the most efficient player types, showing higher efficiency than human players in these combinations.} Finally, although CUAs can achieve human-level success rates, they take much longer to complete the game than humans. This suggests that, under the same experimental conditions, CUAs do not yet use temporal information as efficiently as humans for social coordination.

 % 14 of the 17 most efficient LLM player-partner combinations involve Bayesian players.

\subsection{RQ2: Cognitive Strategies Used by Humans, CUAs, and LLMs}
\label{sec:strategy}

Next, we ask which strategies best capture human, CUA, and LLM behavior.\change{\footnote{For the human experiments, we also collected self-reports. Details can be found in \appref{app:self_report}.}} We evaluated the fit of six types of probabilistic models to each game record: Random, Constant Wait,\footnote{The Constant Wait model assumes that Player A’s waiting time is drawn from the same fitted distribution regardless of the cards played or the preceding game history. This strategy was not included as a rule-based partner condition, but we included it as a potential strategy explaining player behaviors.} Linear +/-Calibration, and Bayesian +/-Calibration.
\begin{comment}
For the Linear models, we evaluated every combination of the initial tempo parameter
$\tau\in\{0.1,0.3,0.5,0.7,1.0,1.3,1.5\}$ and, when calibration was used, the
calibration rate $\eta\in\{0.1,0.3,0.5,0.7,0.9\}$. The Bayesian models used the
same $\tau$ and $\eta$ values and additionally varied the Weibull shape parameter
$\gamma\in\{0.5,0.7,1.0,1.3,1.5,2.0,3.0\}$. For each candidate parameter
combination, the remaining timing-noise or play-rate parameters were fitted by
maximum likelihood, and we kept the combination with the highest likelihood
for that game record. The Random model's play probability and the Constant Wait
model's mean and Gamma shape were fitted directly by maximum likelihood. 
\end{comment}
% For the LLM analysis, we used the condition in which models decided whether to play or wait at one-second intervals. We applied the same fitting procedure to the browser-based interaction records produced by the computer-use agents. 
% Each model assigned a likelihood to the Player A’s sequence of timing decisions. When Player A played first, we calculated the model’s probability that Player A would play within the observed time interval. When Player B played first, we calculated the model’s probability that Player A would still be waiting at that time. Forced autoplay and stand-alone cards were excluded because they did not reflect an active timing decision. The complete likelihood definitions are given in \appref{sec:likelihood}.
We evaluated a wide range of parameters for each model class, and used each model $M$ to compute the likelihood of the critical player's (i.e., human, CUA, or LLM) sequence of actions. When the critical player played first, we calculated the probability under $M$ that the player would play within the observed time interval. When the rule-based partner played first, we calculated the probability under $M$ that the critical player would still be waiting at that time. Forced autoplay and stand-alone cards were excluded because they did not reflect an active timing decision. The complete likelihood definitions are given in \appref{sec:likelihood}.

Using the maximized likelihood for each model and game record, we computed $\mathrm{AIC} = 2K - 2\log \hat{L}$,
% \begin{equation}
%     \mathrm{AIC} = 2K - 2\log \hat{L},
% \end{equation}
where $K$ is the number of fitted parameters and $\hat{L}$ is the maximized likelihood.\footnote{We also considered BIC, which further penalizes model complexity and would therefore disfavor the Bayesian strategies. BIC selected the same model class as AIC for $80.3\%$ of game records. See \appref{bic}.} We selected the model with the lowest AIC for each game record, and then report the proportion of game records won by each model class. AIC values were never compared across different game records. 

\begin{figure}
    \centering
    \includegraphics[width=\linewidth]%{figs/final_alltogether_aic_condensed.png}
    {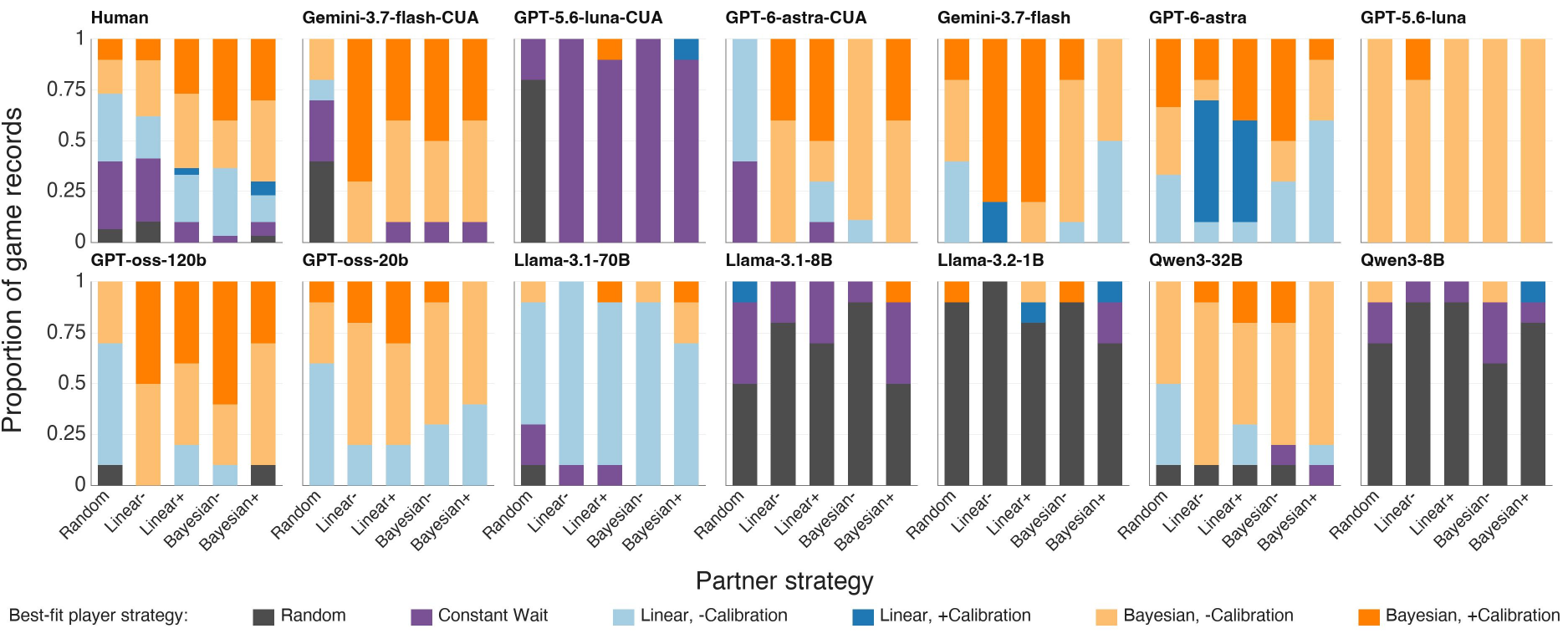}
    % \caption{Proportion of Human, LLM, and CUA Game Records Best Fit by Strategy Models (AIC)}
    \caption{Strategies that best capture human, CUA, and LLM behaviors across partner conditions.} 
    % The x-axis shows rule-based partner types, and the hues show the type of strategy that achieves lowest AIC when fit to the critical player's behaviors.}
    \label{alltogether_aic}
\end{figure}

The results are shown in \Cref{alltogether_aic}. First, we find that humans adopt strategies more flexibly when playing with different partners. The Constant strategy best fit humans paired with simpler partners like Random or Linear players without calibration, while Bayesian strategies best fit humans paired with other more complex partners. In contrast, both LLMs and CUAs tend to be best fit by one strategy across player types. Second, the best fitting strategy for LLM and CUA behavior varies substantially across model families and sizes. Bayesian strategies best fit the behavior for larger models, while the Random strategy best fit the behavior for smaller ones. \texttt{Llama 70B} and \texttt{GPT-5.6-luna (CUA)}, however, are best fit with Linear and Constant strategies, respectively.

%\begin{figure}
%    \centering
%    \includegraphics[width=\linewidth]%{figs/final_humain_aic.png}
%    \caption{Proportion of Human Game Records Best Fit by Each Strategy Model (AIC)}
%    \label{final_human_aic}
%\end{figure}
%\begin{figure}
%    \centering
%    \includegraphics[width=\linewidth]{figs/final_llm_aic.png}
%    \caption{Proportion of LLM Game Records Best Fit by Each Strategy Model (AIC)}
%    \label{final_llm_aic}
%\end{figure}
%\begin{figure}
%    \centering
%    \includegraphics[width=\linewidth]{figs/final_cua_aic.png}
%    \caption{Proportion of Computer-Use Agent Game Records Best Fit by Each Strategy Model (AIC)}
%    \label{final_cua_aic}
%\end{figure}

% \paragraph{Do LLMs have a consistent decision based on time?}
% \begin{itemize}
%     \item Figure~\ref{fig:time}.
%     \item When prompted with different time units, LLMs spend different amount of time for the game. 
% \end{itemize}

% \paragraph{Do LLMs show similar behaviors as humans?}
% \begin{itemize}
%     \item Figures~\ref{fig:time},\ref{fig:success}
%     \item LLMs make more mistakes than humans, but they show similar time usage when playing with different conditions as humans. 
%     \item The larger the model, the better the performance.
%     \item Larger models tend to wait longer.
% \end{itemize}
\subsection{RQ3: Behavior Adaptation in Humans, CUAs, and LLMs}
\label{sec:convention}

Next, we ask whether players adapt to their partners to improve efficiency across the levels of a game. 
% Previous literature on convention formation has primarily used measures such as utterance length and the number of turns to examine the emergence of conventions \citep{clark1986referring,ji2025ad}. As our experiment prohibits communication, these measures are not applicable. Instead, 
Mirroring the analysis in \Cref{sec:overall}, we use two behavioral measures to capture changes of efficiency: normalized hesitation time and success rate.\footnote{Raw hesitation time is not comparable across plays because it depends on the gap between the current card and the previously played card. For example, for the same current card (e.g., 55), a player would be expected to wait for different amounts of time when the previously played card is 12 versus 42. We therefore normalize hesitation time by the value gap between the current and previously played cards.} Because cards that are automatically played following an error by the other partner and standalone cards do not meaningfully reflect the player’s own timing decision, we exclude these cards from the calculation of hesitation time. 

To examine how these measures change over the course of the game, we divide the 110 cards into 10 deciles and model the effect of decile on success rate and normalized hesitation time. For success rate, we fit generalized linear mixed-effects models (GLMMs).\footnote{The model formulas were $\texttt{outcome} \sim \texttt{decile} + (1|\texttt{hand\_id}) + (1|\texttt{group})$, where $\texttt{group}=\texttt{participant}$ for humans and $\texttt{prompt\_type}$ for LLMs and CUAs; $\texttt{outcome}$ was $\texttt{cbind(success, total-success)}$ or $\texttt{normalized\_hesitation\_time}$, respectively.} A decrease in hesitation time suggests that players become more confident in their partner’s strategy, while an increase in success rate suggests that the coordination becomes more successful over time. Neither measure alone, however, provides a complete picture of behavior adaptation: players may become more cautious over time and achieve higher accuracy, or they may simply become less patient and act more quickly. 

\begin{figure}
    \centering
    \includegraphics[width=\linewidth]{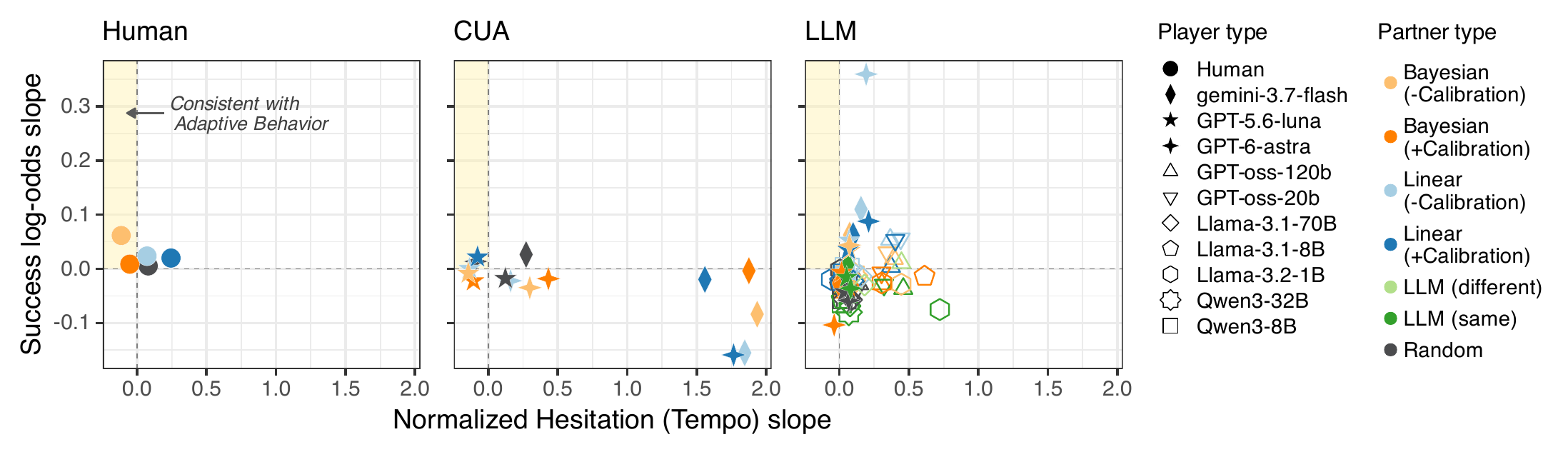}
      \caption{Regression slopes of success rate and normalized hesitation time.}
    \label{fig:alpha_vs_success}
\end{figure}

We plot the slopes of success rate and normalized hesitation time in \Cref{fig:alpha_vs_success}. We find that human players show better learning results across all partner conditions with two conditions showing statistically significant or marginally significant improvements in performance: \LinearNoCal\ condition ($\hat{\beta}=0.024$, $p=0.07$) and \BayesianNoCal\ condition ($\hat{\beta}=0.062$, $p<0.01$). Humans also tend to respond more quickly as the game progresses when playing with Bayesian partners, with negative estimates in both the \BayesianNoCal\ ($\hat{\beta}=-0.11$, $p=0.08$) and \BayesianCal\ ($\hat{\beta}=-0.05$, $p=0.35$) conditions.

In contrast, across all CUA and LLM players, only \texttt{GPT-5.6-luna (CUA)} show both faster speed and higher success rate over the course of the game.\footnote{We do not interpret the \texttt{GPT-6 (CUA)}-Random combination as evidence of convention formation, since Random partners have no temporal structure for a player to adapt to.} However, neither effect is statistically significant ($\hat{\beta}{_\mathrm{tempo}}=-0.08$, $p=0.64$; $\hat{\beta}{_\mathrm{success}}=0.02$, $p=0.30$). Most other player-partner combinations instead exhibit a trade-off between success rate and speed: models either become more successful but slower over the course of the game, or become faster at the cost of a lower success rate. Human participants show patterns consistent with convention formation in two of their five partner conditions, whereas no individual model shows such a pattern in more than one of its five partner conditions. Additionally, human players are the \textbf{only} group that consistently shows increasing success rates over the course of the game across 5 partner types. This pattern suggests that humans adapt to their partners' behavior more consistently across conditions than individual LLMs or CUAs. Details of the statistical results are in \Cref{app:stats}.

\change{To further distinguish player adaptation from partner calibration in the +Calibration conditions, we analyzed errors separately for humans, \texttt{GPT-6}, and their rule-based partners, attributing each error to the player who played too early. We compared error rates in the first and last thirds of each player's 55 cards using paired $t$-tests across games.

Human error rates decreased with both Linear (18.8\% $\rightarrow$ 12.6\%; $t(29) = 2.73$, $p = .011$) and Bayesian partners (14.6\% $\rightarrow$ 10.7\%; $t(29) = 2.60$, $p = .014$), while partner error rates changed by less than 1.5 percentage points ($p > .49$). These results suggest that improvement primarily reflects human adaptation. By contrast, \texttt{GPT-6}'s error rate remained less changed with the Linear partner (9.5\% $\rightarrow$ 9.4\%) and Bayesian partner (10.0\% $\rightarrow$ 7.2\%), suggesting limited adaptation by \texttt{GPT-6}.}

\subsection{RQ4: Behavior Consistency across Prompts (LLMs only)}
\label{sec:time}

% Finally, we evaluated whether the text-prompted LLMs show consistent behavior across different ways of encoding temporal information in the prompt. An agent with a consistent interpretation of time should exhibit similar hesitation times when the temporal granularity or temporal information changes (and all other aspects of the prompt remain identical). However, we find that models are more sensitive to how frequently waiting events occur than to the temporal granularity used to specify them. We use the coefficient of variation (CV) to measure the variation of each player–partner combination across four time-based prompts (\Cref{fig:cv} in \appref{app:time}). In \Cref{fig:time}, we show the hesitation times of the two models with the smallest and largest average CVs (other model results can be found in \Cref{fig:time_other}). We find that simply changing the temporal granularity used to represent waiting time in the prompt can lead to substantially different behavioral patterns, even when all other aspects of the game remain unchanged. This effect is observed consistently across all LLMs. However, its magnitude varies systematically with model size: larger models tend to show smaller CVs, and thus more robust toward prompt variations, than smaller ones.

\begin{figure}[t]
    \centering
    \begin{subfigure}{0.44\textwidth}
        \centering
        \includegraphics[width=\textwidth]{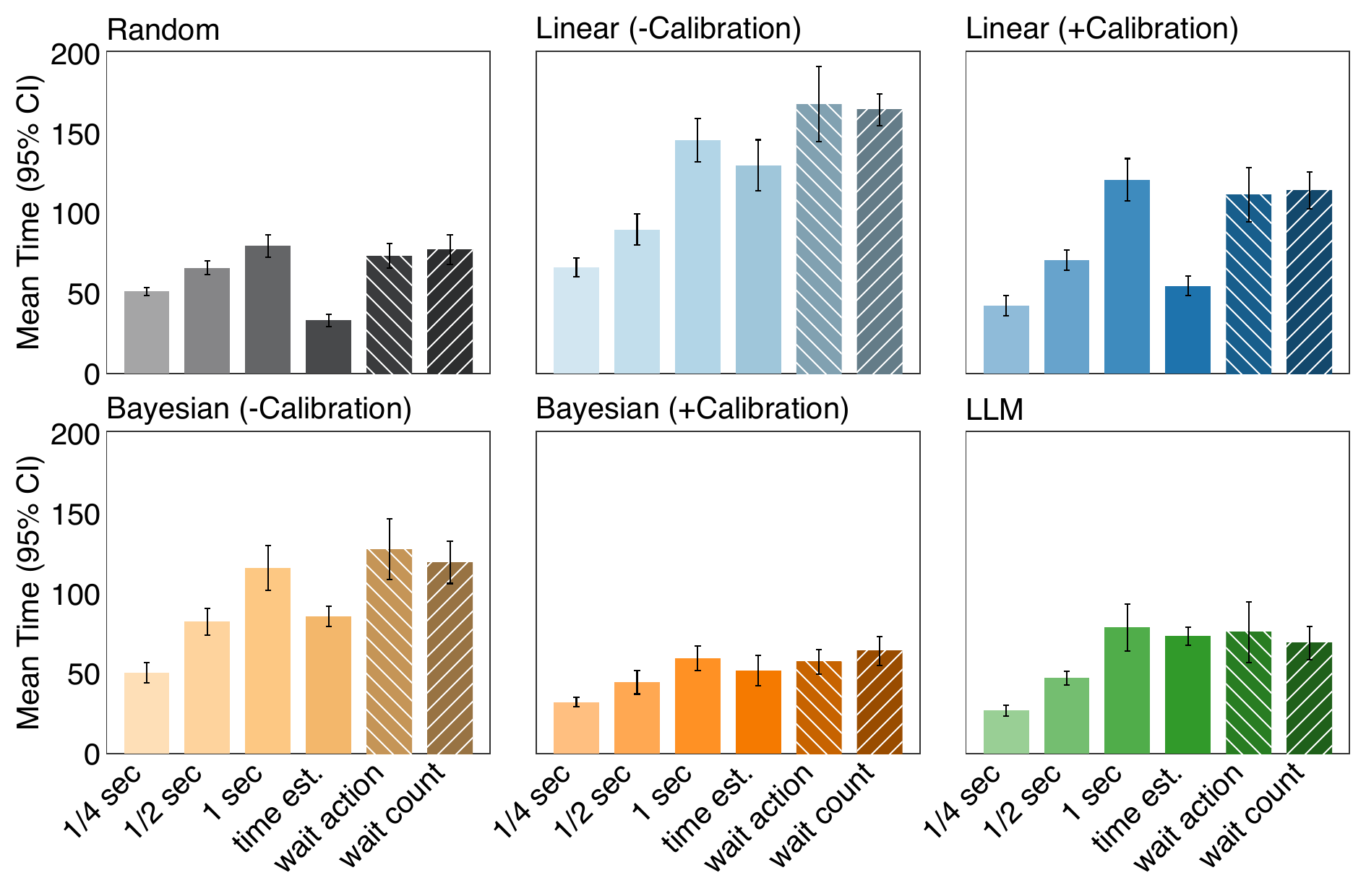}
        \caption{GPT-5.6-Luna}
        \label{fig:time-gpt56}
    \end{subfigure}
    \qquad \qquad
    \begin{subfigure}{0.44\textwidth}
        \centering
        \includegraphics[width=\textwidth]{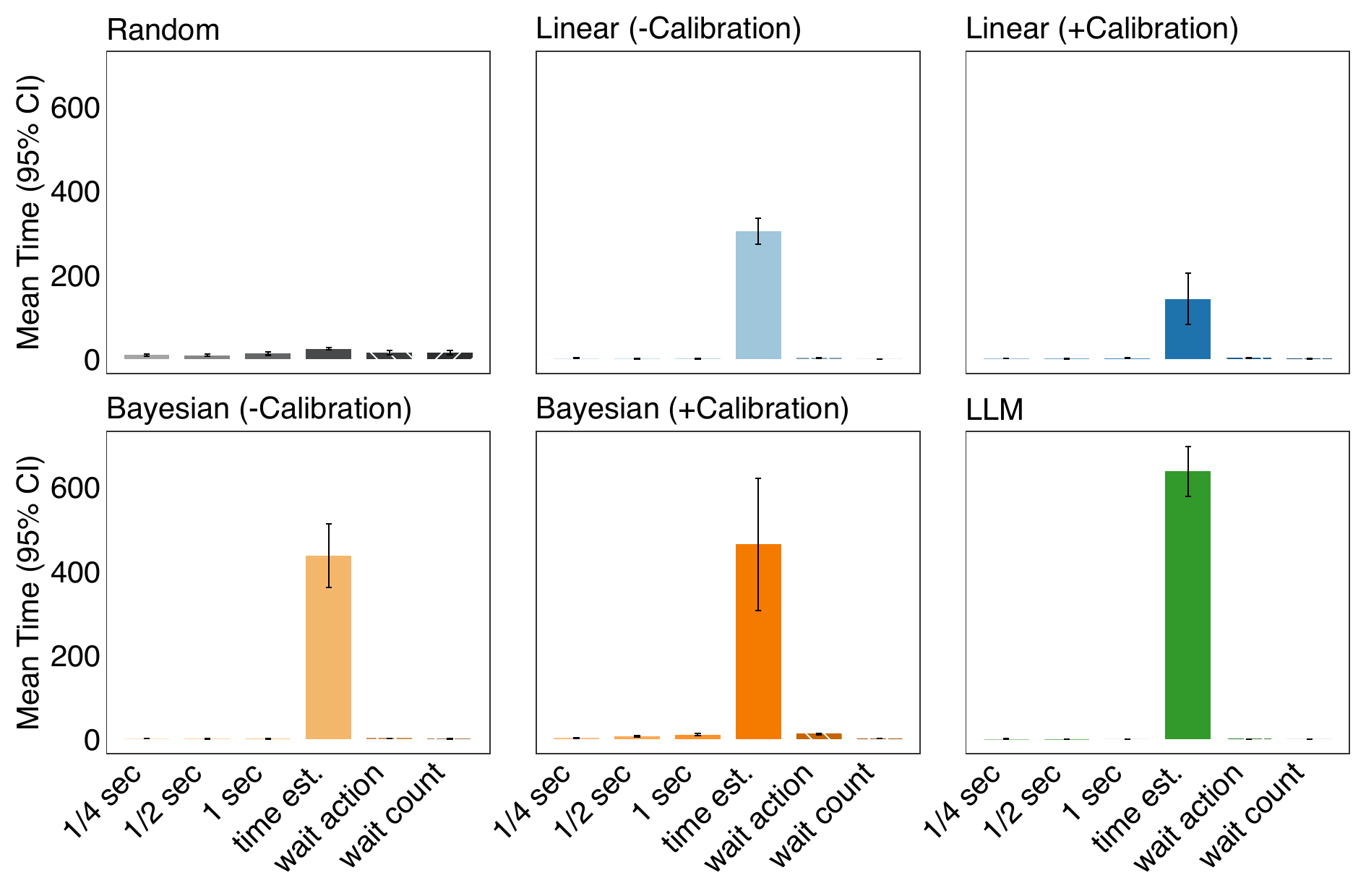}
        \caption{Llama-3.2-1B-Instruct}
        \label{fig:time-llama1b}
    \end{subfigure}

    \caption{Mean play time by partner type and prompt type (x-axis) for the LLMs with the smallest (a) and largest (b) average coefficient of variation across prompt types. Error bars show 95\% CI.}
    \label{fig:time}
\end{figure}

Finally, we evaluated whether the text-prompted LLMs show consistent behavior across different ways of encoding temporal information in the prompt. We first analyze the four Time-based prompts (see \Cref{sec:expt-llm} for details). An agent with a consistent interpretation of time should exhibit similar hesitation times when the temporal granularity or temporal information changes (and all other aspects of the prompt remain identical). 
% However, we find that models are more sensitive to how frequently waiting events occur than to the temporal granularity used to specify them. 
We use the coefficient of variation (CV) to measure the variation of each player–partner combination across the four Time-based prompt variants. \Cref{fig:time} shows the hesitation times of the two models with the smallest and largest average CVs (see \appref{app:time} for additional models and CV results). We find that simply changing the temporal granularity used to represent waiting time in the prompt can lead to substantially different behaviors, even when all other aspects of the game remain unchanged. In particular, it seems that models may make decisions based on how many times they are prompted to wait, rather than \emph{how long} they've been waiting. This effect is observed consistently across LLMs, but its magnitude varies systematically with model size: larger models tend to be more robust to prompt variations.
Turning to the Wait-based prompts, we also observe in some cases (such as with \texttt{GPT-5.6-luna}; \Cref{fig:time-gpt56}) that the Wait-based prompts lead to similar playing time as certain Time-based prompts. Since the Wait-based prompts do not make any reference to time units, this finding suggests that LLMs might be able to perform the task based on numerical information instead of reasoning about real-time.

\change{\section{Discussion \& Conclusion}}

Social interactions in the real world unfold in \emph{time}. Whether and how different agents use time as a signal is therefore central to understanding social coordination for both humans and AI. In this work, we investigated how humans and AI systems reason about the passage of time, extending prior work on one-shot inferences to a collaborative task involving repeated interactions. % by using \themind as a test bed.
%and assess how humans and AI systems cooperate on implicit passage of time. 
% We provide the first empirical evidence on how humans infer and adapt to a partner's behavior from time over the course of repeated interactions. 
We found that humans perform the best with partners that perform Bayesian reasoning about time, and humans themselves also seem to adopt Bayesian reasoning strategies instead of simply mapping between time units and card units. Moreover, humans show evidence of adaptation over the course of the game when playing with Bayesian partners. Taken together, these results suggest that humans make rational use of temporal signals in dynamic interactions, and expect social partners to do the same.

Turning to the AI models, we found that some frontier models can achieve human-level performance and, like humans, show behaviors that are best captured by Bayesian reasoning strategies. However, unlike humans, the models do not seem to adapt to their partner's behavior as the game progresses. Some actually show worse success rates over the course of a game. The text-based LLMs are also sensitive to time granularity presented in the prompts, and show similar timing patterns whether or not time units are mentioned in the prompt. These findings suggest that while AI models might adopt human-like strategies for temporal coordination, the process by which they arrive at this behavior may differ fundamentally from how humans learn to coordinate with a partner. 

We also find that only prompt-based LLMs achieve human-comparable efficiency in the task, even though CUAs act in real-time like humans do. %that is, when the passage of time is explicitly represented in the prompt. 
% As shown in \Cref{fig:alpha_vs_success}, the two SOTA CUAs (\texttt{GPT-6 (CUA)} and \texttt{Gemini (CUA)}) exhibit increasing waiting times and declining success rates as the game progresses.
%Although CUAs match or even exceed humans in success rate, they take longer to complete the game (\Cref{fig:alpha_vs_success}). 
This pattern cannot be explained simply by implementation or reasoning latency, since CUAs do show hesitation times comparable to humans when playing against certain kinds of partners. This observation may instead reflect a difficulty current CUAs face in balancing speed and accuracy as an interaction unfolds, trading longer waiting times for higher accuracy.

\change{To our knowledge, our study is the first to systematically study how humans and AI models use time---and no other communication---to achieve complex cooperative goals. 
% By comparing humans, LLMs, and CUAs on \themind, we see our study as suggesting 
More broadly, our work suggests new ways to investigate the flexibility and efficiency of social cognition in humans and artificial systems when coordination relies on implicit signals.}

\change{\section{Limitations \& Future Work}}
We note a few limitations regarding our study. First, \change{as mentioned in \Cref{sec:expt-human}}, we focused on players interacting with rule-based partners, rather than directly pairing ``live'' players with each other \change{(with the exception of LLMs paired with LLMs)}. \change{This choice enables a finer-grained analysis of strategies used by the players and comparison across different types of players.} While this design gave us greater control for addressing our research questions, it does not tell us what kinds of strategies would organically arise through player-player interactions. \change{Investigating how humans would naturally play the game with each other is an exciting direction for future work.} 
% Second, cost: ensuring data quality in human–human pairs requires excluding an entire session whenever either player fails an attention check, substantially increasing the number of sessions needed; frontier CUA models are also costly to run at scale, making CUA-CUA and Human-CUA pairs similarly expensive to collect at a comparable sample size. We leave these more complex player-player interactions to future work. 

Second, the strategies we use to model the players' behavior are necessarily simplified. Players may mix strategies, change strategies across time, and so on. \change{Other factors, such as the card value or player's personal sense of timing, can also influence how player plays the card. }Additionally, the Bayesian strategy we used in our experiment represents a \change{simple} rational approach to inferring the state of the game, but a \change{more sophisticated} social strategy might involve recursive reasoning about the other agent's mental state\change{s \citep[e.g.,][]{frank_predicting_2012,baker_rational_2017}}. We leave a detailed treatment of these more complex strategies for future work. 

Third, we note a key difference in the way the task is realized for humans and text-prompted LLMs: humans experience continuous wall-clock time, whereas for LLMs we approximate continuous-time interaction by explicitly specifying waiting durations in prompts. While this distinction is motivated by our broader question---how a non-time-embodied system can perform the task---we note that this complicates direct comparisons between human and LLM playing times. A more direct comparison would test humans in a text-based version of the task\change{, although this differs drastically from how humans interact in natural settings.}

Finally, our experiments use a constrained task setting \change{where temporal signals are fully isolated from other forms of communication}. \change{This task is ideal for addressing our research questions because it tests the limits of humans' and models' ability to reason about time: there is no other channel for information exchange to complete the task. It remains an open question how our findings would extend to settings where} temporal signals interact with \change{language and other communicative} cues.
% in more naturalistic settings.  \change{Our findings concern temporal coordination in a setting with delayed but verifiable outcomes. Players interpret their partner's silence before receiving feedback, much as conversational inferences from pauses may be confirmed or repaired in later turns. Our findings may therefore extend to other interactions with delayed feedback. However, many social interactions lack clear outcome feedback; for example, one may infer reluctance from a friend's delayed reply to a message without ever learning whether the inference was correct. Whether and how our findings extend to such settings remains an open question.} 
% Finally, our experiment design deliberately isolate temporal signals to provide a controlled test of their use. In naturalistic settings, timing can be entangled by other noises such as participants' perception of time and communication purposes. Future work can examine whether our findings generalize to such richer settings.

% \todo{might be nice to connect to the notion of timescapes: \citet{singhal_timescapes_2026} I think this could help differentiate what we're doing from standard temporal reasoning benchmarks.}

% Hence, to build reliable computational assistant that interacts with humans or agents in real-time circumstances, we need more careful evaluation, and revisit AI architecture.

% \jh{if space allows, would be nice to end with something connecting to bigger picture, kind of like the last sentence of the abstract}\todo{cogsci: AI:}

\subsection*{Acknowledgments}

We thank Michael Franke and lab members from the Group for Language \& Intelligence and NERT for their helpful feedback. J.H.~gratefully acknowledges support from Google under a Gemini Academic Program Award. 

\subsection*{AI use statement}

In this work, we used generative AI tools for \textit{design or provide feedback on research  methodology or experiments, interpret results}.
We have not used generative AI tools for \textit{implement methods, clean and reformat dataset, support qualitative and thematic data analysis, propose or refine hypotheses},
and all other tasks are not applicable to this work.
Additionally, we used generative AI tools for \textit{create or modify scientific figures or images, sourcing/searching for information, edit a research paper to improve readability, identify relevant literature}. We have reviewed all AI-assisted work. The implementation python code is written by authors and then reviewed by AI tools. The JavaScript implementation of the game interface was translated from the Python code with the assistance of AI tools and was tested by the authors before the experiments. We also used AI tools to search for information and identify relevant literature; all such information was independently verified against the original sources.  We take responsibility for the final content of this work,
including text, claims or artifacts produced with the aid of generative AI.

\subsection*{Ethics statement}
This study involved human participants and was reviewed and approved by an institutional review board. All participants provided informed consent before participating. Participants were compensated for their time in accordance with the approved study protocol (\$12 per hour). The data were collected and analyzed under the approved protocol, and personally identifying information was not included in the released or analyzed dataset.

Our work focuses on how humans and AI coordinate with a rule-based partner by using implicit signals, i.e., passage of time, through repeated interactions. As AI systems increasingly serve as personal assistants and collaborative agents, the ability to infer information from behavioral signals may help build more efficient and personalized interactions. However, it is also worth noting that such ability may lead to unintended social inference and raise privacy concerns, particularly during interactions in which users may not be fully aware of such subtle cues. 

\subsection*{Reproducibility statement}
The hyperparameters used by the models are listed in \Cref{tab:hyperparameters}. 

\change{Code is available: \url{https://github.com/hulaboratory/the-mind}}.

\subsubsection*{Author Contributions}
X.Y.~is responsible for the experiment design, implementation, data collection, and analysis of the human, LLM, and CUA experiments (RQs 1,3,4), as well as the paper framing and writing.

S.G. is responsible for developing the mathematical framework jointly with J.H. (including the theoretical framing, formal strategies, and appendix derivations), conducting the AIC model-comparison analyses, and writing the corresponding sections. 

J.H.~is responsible for conceptualization, experiment design, paper writing and editing, providing resources for computational and human experiments, and supervising the project.

\bibliography{iclr2027_conference}
\bibliographystyle{iclr2027_conference}

\appendix

\section{Details of rule-based players}
\label{app:players}

%% modified: added a sentence here so the appendix's conditioning shorthand is defined.

% , and
% we abbreviate conditioning on the observed history
% $\mathbf{C}^{(t)} = \mathbf{c}^{(t)}$ as conditioning on $\mathbf{c}^{(t)}$. The tempo parameter $\tempo$ is defined in the main text; its calibration rule
% is derived in Section~\ref{app:tempo}.

\subsection{State Space}

Throughout this appendix we use the notation introduced in \Cref{sec:rule-based}. There are two players, A and B. At the start of level $n$, each player is dealt a hand of $n$ cards uniformly at random from a deck $\{1, 2, \dots, 100\}$. 
% We write $c^{(t)}$ to denote the most recently played card at time $t$,
% and $\mathbf{c}^{(t)} = \{c_1, c_2, \dots, c^{(t)}\}$ to denote the history of cards played until time $t$.
% We also write $a_1^{(t)}$ to denote player A's lowest remaining card, and $b_1^{(t)}$ to denote player B's lowest remaining card at time $t$. The $t$ ``clock'' is reset to 0 at the start of each round and each time a card is played.
We denote the players' hands as the following:
\begin{align}
    \text{Player A holds } \mathbf{A} &= \langle a_1, a_2, \cdots, a_n\rangle, \text{ where } a_1 < a_2 < \dots < a_n \\
    \text{Player B holds } \mathbf{B} &= \langle b_1, b_2, \cdots, b_n\rangle, \text{ where } b_1 < b_2 < \dots < b_n
\end{align}
We will denote $a_1^{(t)}$ as player A's lowest remaining card, 
$b_1^{(t)}$ as player B's lowest remaining card, and $c^{(t)}$ the most recently played card at time $t$. When no card has yet been played in the current level, we set $c^{(t)} = 0$. There is no 0 card in the deck.

At any given time $t$, we can define the state as:
\begin{itemize}
    \item $\mathbf{A}^{(t)} \subseteq \mathbf{A}$ denotes player A's remaining cards (known only to A)
    \item $\mathbf{B}^{(t)} \subseteq \mathbf{B}$ denotes player B's remaining cards (known only to B)
    \item $\mathbf{c}^{(t)} = \{c_1, c_2, \ldots, c^{(t)}\}$ is the history of played cards (known to both players)
    \item $\overline{\mathbf{c}}^{(t)} = \{1, \ldots, 100\} \setminus \mathbf{c}^{(t)}$ is the set of cards not yet played (known to both players, since all played cards $\mathbf{c}^{(t)}$ are known)
\end{itemize}

\subsection{Linear player}
\label{app:counting}
% The core intuition is that the game requires players to somehow form a
% ``tempo'' convention, without explicitly communicating. We will denote this
% tempo with the variable $\tempo > 0$: a smaller $\tempo$ means that everyone plays
% more quickly; a larger $\tempo$ means that everyone plays more slowly. Assume that $\tempo$ is fixed (i.e., determined according to one of the rules
% from Section~\ref{app:tempo}). Player A simply waits a fixed amount of time
% proportional to the gap between their lowest card and the previously played
% card before playing:
As described in \Cref{sec:strategies}, the Linear player simply waits a fixed amount of time
proportional to the gap between their lowest card and the previously played
card before playing. We denote this scaling factor as with a ``tempo'' parameter $\tempo > 0$.

Therefore, player A would play their lowest card at time
\begin{equation}
    t^* = \tempo \cdot \bigl(a_1^{(t)} - c^{(t)}\bigr),
    % + \varepsilon_A, \quad \varepsilon_A \sim \mathcal{N}(0, \sigma_A^2)
\end{equation}
and similarly for player B. When a single player holds the only card(s) remaining in the level (i.e., the standalone cards), they can execute the rest of the plays perfectly, so they play the remaining card(s) immediately rather than waiting the full time.

If both players share the same \tempo and follow this strategy perfectly and deterministically, this strategy will never make mistakes. However, as stated in \Cref{sec:strategies}, this strategy ignores the \emph{information} in the other player's silence, and is sensitive to miscalibration of $\tempo$. For Linear players with tempo calibration (see \Cref{sec:tempo-calibration}), we replace \tempo by the updated estimate after each play.

\subsection{Bayesian player}
\label{app:bayesian}

Consider the game state at time $t$. Let $m_A$ denote the number of cards that player A still holds
($|\mathbf{A}^{(t)}|$), and let $m_B$ denote the number of cards that player B
still holds ($|\mathbf{B}^{(t)}|$). Both players know $m_A$ and $m_B$ at all times.
% After $x = |\mathbf{c}^{(t)}|$ total cards
% have been played on the table, $m_A + m_B = 2n - x$, but the individual values
% depend on \emph{who} played each of those $x$ cards. Since all played cards
% are visible, both players can track this exactly. Player A knows how many of
% the played cards were theirs, so they can infer $m_B = 2n - x - m_A$ (and
% vice versa).

\paragraph{The prior.} After some cards
$\mathbf{c}^{(t)}$ have been played, player A knows their own remaining hand
$\mathbf{A}^{(t)}$, and knows that player B's remaining cards
$\mathbf{B}^{(t)}$ are drawn from
$\overline{\mathbf{c}}^{(t)} \setminus \mathbf{A}^{(t)}$.

The prior probability that player B's current lowest card $B_1^{(t)}$ is some card value $k$ is the probability that $k$ is the minimum of $m_B$ cards drawn uniformly
without replacement from a set of
$N = |\overline{\mathbf{c}}^{(t)} \setminus \mathbf{A}^{(t)}|$ cards. For this
to be true, two conditions must be satisfied: (1) $k$ needs to be in B's hand, and (2) all remaining $m_B - 1$ of B's cards are greater than $k$. This gives us the following formula for the history-conditioned
\textbf{prior} over player B's lowest remaining card
$B_1^{(t)}$:
\begin{equation}
    P(B_1^{(t)} = k \mid \mathbf{c}^{(t)}) =
    \frac{\binom{|\overline{\mathbf{c}}^{(t)} \setminus \mathbf{A}^{(t)}| - 1 - \mathrm{rank}(k)}{m_B - 1}}
         {\binom{|\overline{\mathbf{c}}^{(t)} \setminus \mathbf{A}^{(t)}|}{m_B}}
    \label{eq:app-prior}
\end{equation}
The denominator is the total number of outcomes (i.e., the number of ways to
choose B's hand of $m_B$ cards from the available cards). The numerator is a bit more involved. Given that $k$ is in player B's hand,
the remaining $m_B - 1$ cards must be chosen from the cards in
$\overline{\mathbf{c}}^{(t)} \setminus \mathbf{A}^{(t)}$ that are $> k$. We
can denote this as $\mathrm{rank}(k)$. So, the number of elements greater
than $k$ is given by
$|\overline{\mathbf{c}}^{(t)} \setminus \mathbf{A}^{(t)}| - 1 - \mathrm{rank}(k)$.
Notice that the revealed card acts as a lower bound on all remaining cards in
that player's hand, which tightens the prior at each step. % As the game
% progresses, both players have increasingly precise beliefs about each other's
% remaining hands, which is why coordination tends to improve within a round as
% cards are played.

\paragraph{Likelihood (survival function).} Player A observes that player B has been silent
for time $t$ since the last card was played. The likelihood of observing silence
up to time $t$, given $B_1^{(t)} = k$, is the survival function of a hazard
process. To capture the intuition that the hazard rate may change over time, we take $(T_B \mid B_1^{(t)} = k, \mathbf{c}^{(t)})$ to be distributed as Weibull with shape $\gamma$ and scale $\tempo \cdot (k - c^{(t)})$, where \tempo is the tempo parameter, and $\gamma$ is a parameter
controlling the change of the hazard rate. If $\gamma=1$, this is equivalent to an Expontential distribution. If $\gamma > 1$, the hazard rate
increases over time, which captures impatience and pressure to act. This gives us the following survival function:
\begin{equation}
    P(T_B > t \mid B_1^{(t)} = k, \mathbf{c}^{(t)}) = e^{-\left(t/(\tempo \cdot (k - c^{(t)}))\right)^{\gamma}} \label{eq:app-gap-survival}
\end{equation}
In our experiments, we set $\gamma = 2$ for the Bayesian rule-based partner, based on feedback from a pilot study.

% We consider the \emph{gap} between the previously
% played card $c^{(t)}$ and player B's current lowest card $B_1^{(t)}$:
% \begin{equation}
%     P(T_B > t \mid B_1^{(t)} = k, \mathbf{c}^{(t)}) = e^{-t/(\tempo \cdot (k - c^{(t)}))}
%     \label{eq:app-gap-survival}
% \end{equation}

%% modified: "In Round 1, c_{i-1} doesn't exist" is modified to use the c^{(t)} = 0 convention.
% At the start of a round, $c^{(t)} = 0$, so we just recover
% Equation~\ref{eq:app-exp-survival}.

%% modified: the "NOTE: we can make tau an explicit function of C" is rewritten as a statement of what calibrated players do.
% Since the survival function depends on $\mathbf{c}^{(t)}$, 
For Bayesian players with tempo calibration (see \Cref{sec:tempo-calibration}) we replace $\tempo$ by the updated estimate after each play, which is itself a function of the previous game history.

\paragraph{Bayesian updating.} At each time point $t$, Player A can update their
belief over $B_1^{(t)}$, the lowest card in player B's remaining hand. By Bayes' rule,
\begin{equation}
    P(B_1^{(t)} = k \mid T_B > t, \mathbf{c}^{(t)}) =
    \frac{P(T_B > t \mid B_1^{(t)} = k, \mathbf{c}^{(t)}) \cdot P(B_1^{(t)} = k \mid \mathbf{c}^{(t)})}
         {P(T_B > t \mid \mathbf{c}^{(t)})}
\end{equation}
where the prior and likelihood are given by \Cref{eq:app-prior} and \Cref{eq:app-gap-survival}, respectively.

% We can write out the denominator by explicitly marginalizing:
% \begin{equation}
%     P(T_B > t \mid \mathbf{c}^{(t)}) =
%     \sum_{j \in \overline{\mathbf{c}}^{(t)} \setminus \mathbf{A}^{(t)}}
%     P(T_B > t \mid B_1^{(t)} = j, \mathbf{c}^{(t)}) \cdot P(B_1^{(t)} = j \mid \mathbf{c}^{(t)})
% \end{equation}
% Substituting and putting everything together, this gives us:
% \begin{equation}
%     P(B_1^{(t)} = k \mid T_B > t, \mathbf{c}^{(t)}) =
%     \frac{e^{-t/(\tempo \cdot (k - c^{(t)}))} \cdot P(B_1^{(t)} = k \mid \mathbf{c}^{(t)})}
%          {\sum_{j \in \overline{\mathbf{c}}^{(t)} \setminus \mathbf{A}^{(t)}}
%           e^{-t/(\tempo \cdot (j - c^{(t)}))} \cdot P(B_1^{(t)} = j \mid \mathbf{c}^{(t)})}
% \end{equation}

\paragraph{Decision rule.} Player A plays their lowest card
($a_1^{(t)}$) at the first time $t^*$ such that they are sufficiently
confident (according to some threshold $\theta$) that $a_1^{(t)} < B_1^{(t)}$.
That is:
\begin{equation}
    t^* = \min\{t \geq 0 : P(B_1^{(t)} > a_1^{(t)} \mid T_B > t, \mathbf{c}^{(t)}) \geq \theta\}
\end{equation}
To compute $P(B_1^{(t)} > a_1^{(t)} \mid T_B > t, \mathbf{c}^{(t)})$, we sum the relevant values of the posterior:
\begin{align}
    P(B_1^{(t)} > a_1^{(t)} \mid T_B > t, \mathbf{c}^{(t)})
    &= \sum_{k \in \overline{\mathbf{c}}^{(t)} \setminus \mathbf{A}^{(t)} \,:\, k > a_1^{(t)}}
       P(B_1^{(t)} = k \mid T_B > t, \mathbf{c}^{(t)}) 
    % &= \sum_{k > a_1^{(t)},\, k \in \overline{\mathbf{c}}^{(t)} \setminus \mathbf{A}^{(t)}}
    %    P(B_1^{(t)} = k \mid T_B > t, \mathbf{c}^{(t)}) 
    % &= \sum_{k > a_1^{(t)},\, k \in \overline{\mathbf{c}}^{(t)} \setminus \mathbf{A}^{(t)}}
    %    \frac{e^{-t/(\tempo \cdot (k - c^{(t)}))} \cdot P(B_1^{(t)} = k \mid \mathbf{c}^{(t)})}
    %         {\sum_{j \in \overline{\mathbf{c}}^{(t)} \setminus \mathbf{A}^{(t)}}
    %          e^{-t/(\tempo \cdot (j - c^{(t)}))} \cdot P(B_1^{(t)} = j \mid \mathbf{c}^{(t)})}
\end{align}

\subsection{Random player}
\label{app:random}
At every second, random player A samples from a Bernoulli distribution with $p=0.5$ to decide whether to play $a_1^{(t)}$ or wait.

\section{Computing Data Likelihood} \label{sec:likelihood}

To compute AIC, we need to compute the likelihood of the plays observed for a single game record.

Let the data be given by a sequence of $D$ datapoints, each of the form $(x_i, c_i)$, where $x_i$ is the play (card and time) and $c_i$ is the context of previous plays. The log-likelihood of the data for a given model with parameters $\theta$ is given by:
\begin{align}
    \sum_{i=1}^D \log P(x_i \mid c_i; \theta)
\end{align}

\subsection{Linear model}

The Linear strategy predicts Player A's waiting time from the gap between
Player A's lowest card and the previously played card. It multiplies this gap by the current tempo parameter. We used a Gamma distribution centered on this predicted waiting time, which allows the actual plays to occur earlier or later, but still maintains nonnegative waiting times. Using the notation defined in the main text, we modeled
\begin{equation}
T_{A,i}
\sim
\operatorname{Gamma}
\left(
    \kappa,\,
    \frac{
        \max\left\{
            \tempo_i\left(a_1^{(t)}-c(t)\right),
            \delta_i/2
        \right\}
    }{\kappa}
\right),
\end{equation}
where the Gamma distribution is parameterized by shape and scale. 

Its mean is
\[
\max\left\{
    \tempo_i\left(a_1^{(t)}-c(t)\right),
    \delta_i/2
\right\},
\]
and its variance is the square of the mean divided by $\kappa$. The parameter $\kappa$ controls timing variability: larger values produce waiting times more tightly concentrated around the Linear prediction.

\begin{comment}
 In the calibration-on case, the tempo estimate was updated after each observed play. In the calibration-off case, it remained fixed. If Player B played first, we calculated the probability that Player A would still be waiting at that time.
\end{comment}

Without calibration, $\tempo_i$ remained fixed throughout the game. With calibration, the tempo estimate was updated after each observed, non-forced play using the rule defined in the main text:
\begin{equation}
    \hat{\tempo}_i
    =
    (1-\eta)\hat{\tempo}_{i-1}
    +
    \eta\frac{t_i}{c_i-c_{i-1}}.
\end{equation}
Here, $\eta$ determines how strongly the most recent observed play changes the current tempo estimate.

\subsection{Bayesian model}
% Turn the decision rule into a distribution?

The Bayesian strategy uses Player B's silence to update Player A's confidence that Player A holds the lower card:
\[
P\left(
    B_1^{(t)}>a_1^{(t)}
    \mid T_B>t,C(t)
\right).
\]
As described earlier, this posterior is based on the Weibull survival model:
\[
P\left(
    T_B>t
    \mid B_1^{(t)}=k,C(t)
\right)
=
\exp\left[
    -\left(
        \frac{t}{\tempo(k-c(t))}
    \right)^\gamma
\right].
\]

Instead of using a fixed decision threshold, we converted Player A's confidence into an instantaneous play rate:
\[
\lambda_A(t)=
\exp\left[
    \beta_0+
    \beta_1
    P\left(
        B_1^{(t)}>a_1^{(t)}
        \mid T_B>t,C(t)
    \right)
\right],
\]
where $\beta_1\geq0$. The parameter $\beta_0$ determines the baseline play rate, and $\beta_1$ determines how strongly the play rate increases with confidence. We do this to give Player A a nonzero play rate even at low confidence. 

The probability that Player A remains waiting until time $t$ is then:
\[
P(T_A>t)
=
\exp\left[
    -\int_0^t\lambda_A(u)\,du
\right].
\]
So, if Player A played during the observed interval $[t_1,t_2)$, we calculated:
\[
P(t_1\leq T_A<t_2)
=
P(T_A>t_1)-P(T_A>t_2).
\]
If Player B played first at time $t$, we instead calculated $P(T_A>t)$, the probability that Player A would still be waiting at that time. In the +Calibration condition, $\tempo$ was updated after each observed play using the calibration rule defined above.

\subsection{Random strategy}
%Exponential distribution?
The Random strategy assumes that Player A has a constant probability $p$ of choosing to play during each one-second interval. Let $N$ be the zero-indexed one-second interval in which Player A plays:
\begin{equation}
P(N=n)=p(1-p)^n,
\qquad n=0,1,2,\ldots.
\end{equation}
Conditional on selecting interval $N$, the play time was uniform within that second:
\begin{equation}
T_A=N+U,
\qquad
U\sim\operatorname{Uniform}(0,1).
\end{equation}
This allowed the model to assign probabilities to continuous and sub-second observations. For $t\geq0$, let $n=\lfloor t\rfloor$ and $r=t-n$. The
resulting CDF is
\begin{equation}
F(t)
=
1-(1-p)^n+p(1-p)^n r.
\end{equation}
The parameter $p$ was fitted separately to each game record. 
%Event-bin and survival probabilities were calculated from this CDF using the common likelihood definition above.

\subsection{Constant strategy}
The Constant wait model assumed that Player A's waiting time did not depend on the card values or game history, so the expected waiting time is constant:
\begin{equation}
    T_i
    \sim
    \operatorname{Gamma}
    \left(\kappa,\frac{\mu}{\kappa}\right),
\end{equation}
where the Gamma distribution is parameterized by shape and scale. It has mean $\mu$ and variance $\mu^2/\kappa$. Both $\mu$ and $\kappa$ were fitted separately to each game record. %Player A plays contributed Gamma event-bin probabilities, and Player B plays contributed Gamma survival probabilities.

\subsection{Parameter fitting and model complexity}

For each candidate model and game record, we selected the parameter values that maximized the likelihood. We evaluated the initial tempo parameter over
\begin{equation}
\tempo
\in
\{0.1,0.3,0.5,0.7,1.0,1.3,1.5\},
\end{equation}
the calibration learning rate over
\begin{equation}
\eta
\in
\{0.1,0.3,0.5,0.7,0.9\},
\end{equation}
and the Weibull shape parameter over
\begin{equation}
\gamma
\in
\{0.5,0.7,1.0,1.3,1.5,2.0,3.0\}.
\end{equation}

The Random probability $p$ was optimized continuously over $(0,1)$. For the Gamma models, $\kappa$ was optimized continuously over
\begin{equation}
\kappa\in[0.02,10{,}000].
\end{equation}
For Constant Wait, $\mu$ was optimized continuously over a positive, record-specific range covering the observed waiting times. For Confidence Bayesian, $\beta_0$ and $\beta_1$ were optimized continuously over
\begin{equation}
\beta_0\in[-12,6],
\qquad
\beta_1\in[0,20].
\end{equation}

Values selected from the parameter grids were counted as fitted parameters. All candidate models were fitted to the same timing decisions within a game record, and values were compared only within that record.

\section{Human Experiments}
\label{human-exp}
All participants were native English speakers residing in the United States. To ensure data quality, we restricted participation to Prolific users with an approval rate above 95\%. To guarantee the data collection quality, we used hesitation time on standalone cards (i.e., the cards played at the end of each level after the partner has finished playing) as an attention check, and filtered out participants whose hesitation time exceeded 1 minute (1 participant was excluded on this basis). Participants were compensated at an hourly rate of \$12. We provide participants a performance-based bonus of up to \$2. We computed a performance score as $-(0.5 \cdot \text{time}_z + 0.5 \cdot \text{error}_z)$, where $\text{time}_z$ and $\text{error}_z$ denote the $z$-scored completion time and error rate, respectively, computed across all participants. Participants in the top 25\% of the performance-score distribution received a \$2 bonus, while those in the 25th-50th percentiles received a \$1 bonus.

\subsection{Game Interface for Human and CUA Experiments}
\label{app:game_layout}
\begin{figure}[!th]
    \centering
    \includegraphics[width=\linewidth]{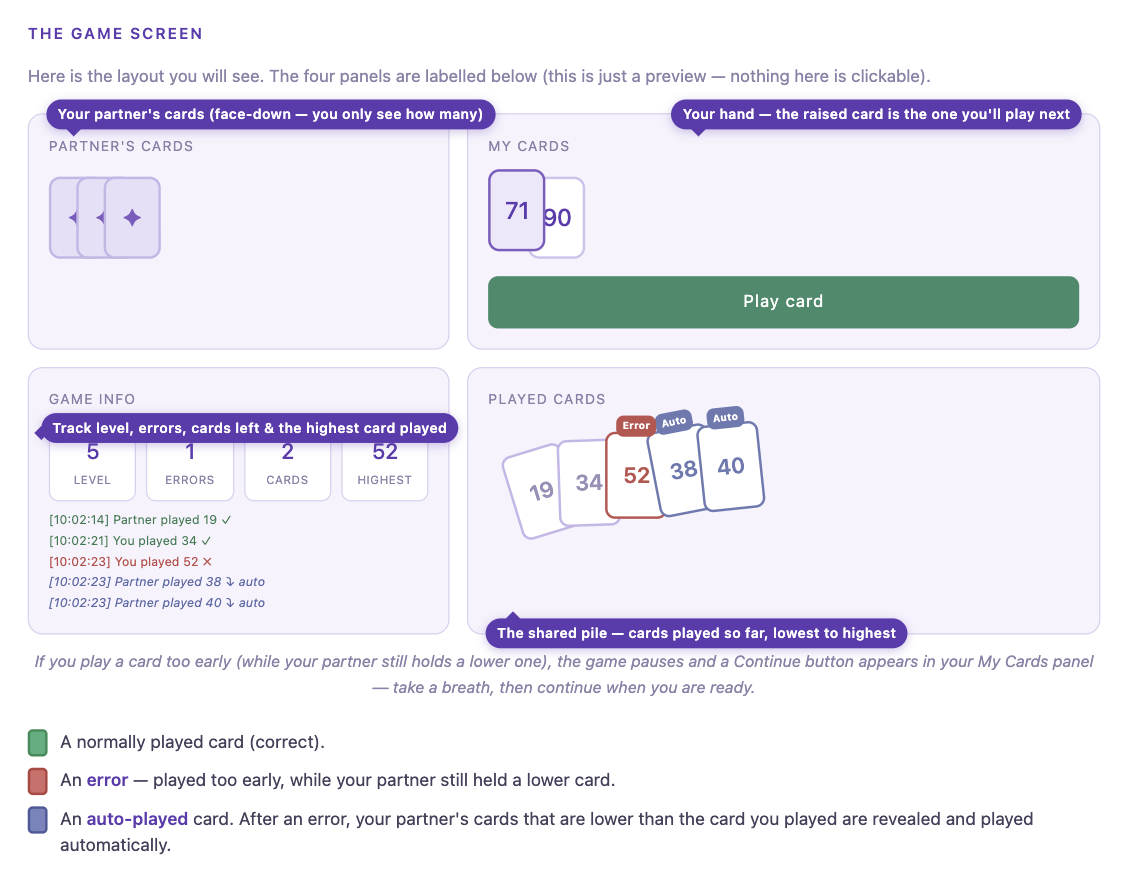}
    \caption{The game interface of \textit{The Mind}.}
    \label{fig:game_interface}
\end{figure}

\begin{figure}
    \centering
    \includegraphics[width=\linewidth]{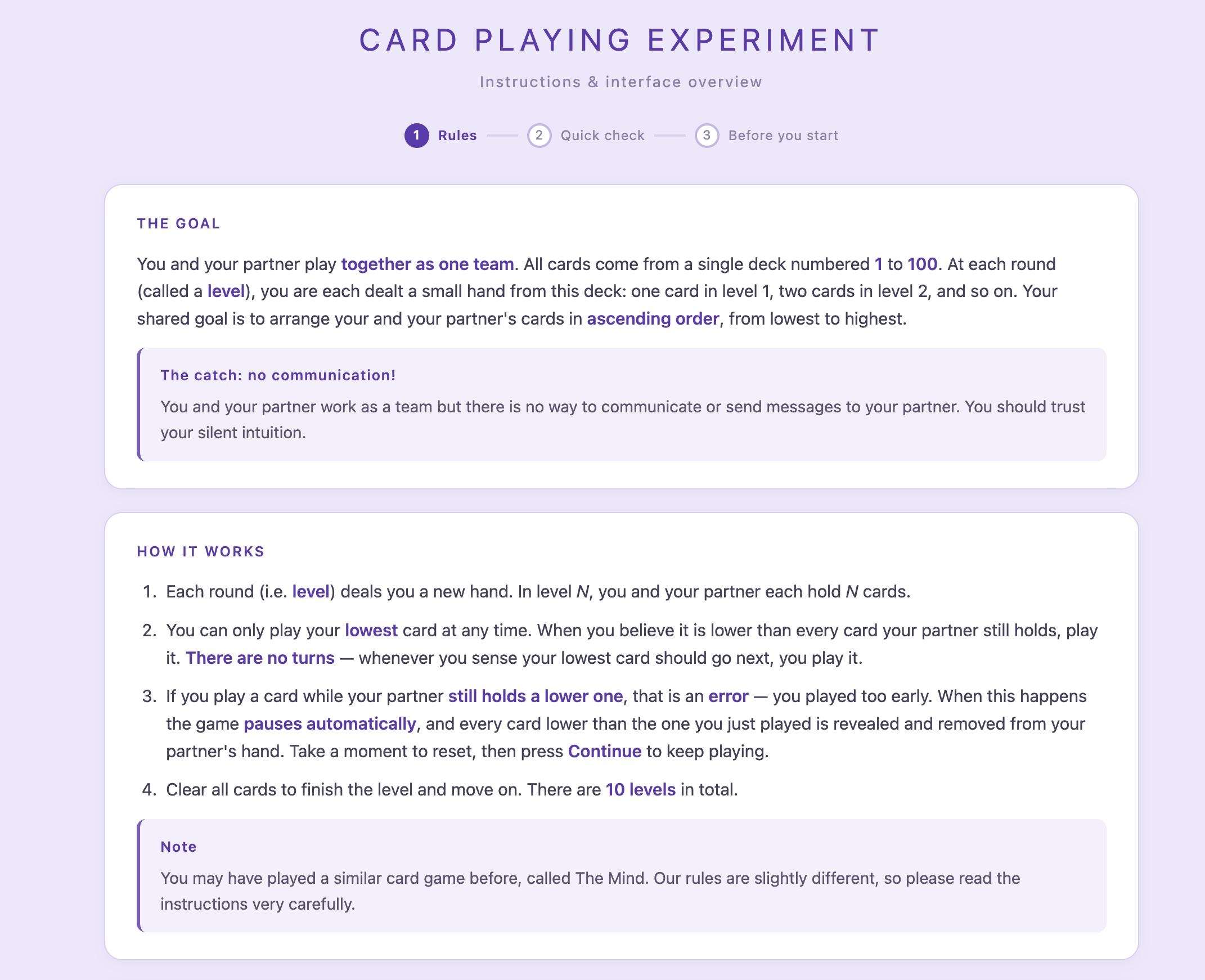}
    \caption{Instruction Page (1/3)}
    \label{fig:ins1}
\end{figure}

\begin{figure}
    \centering
    \includegraphics[width=\linewidth]{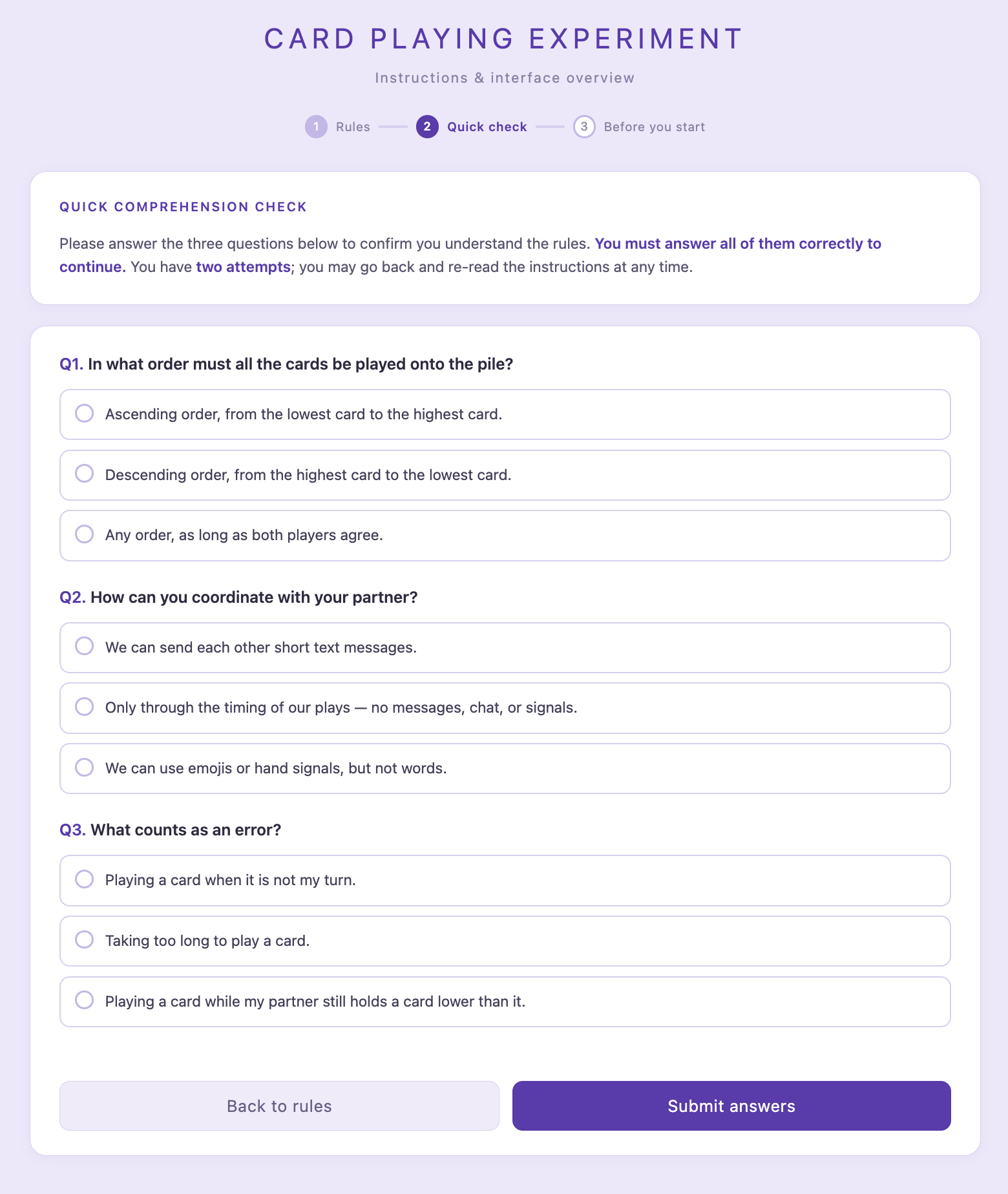}
    \caption{Instruction Page (2/3)}
    \label{fig:ins2}
\end{figure}

\begin{figure}
    \centering
    \includegraphics[width=\linewidth]{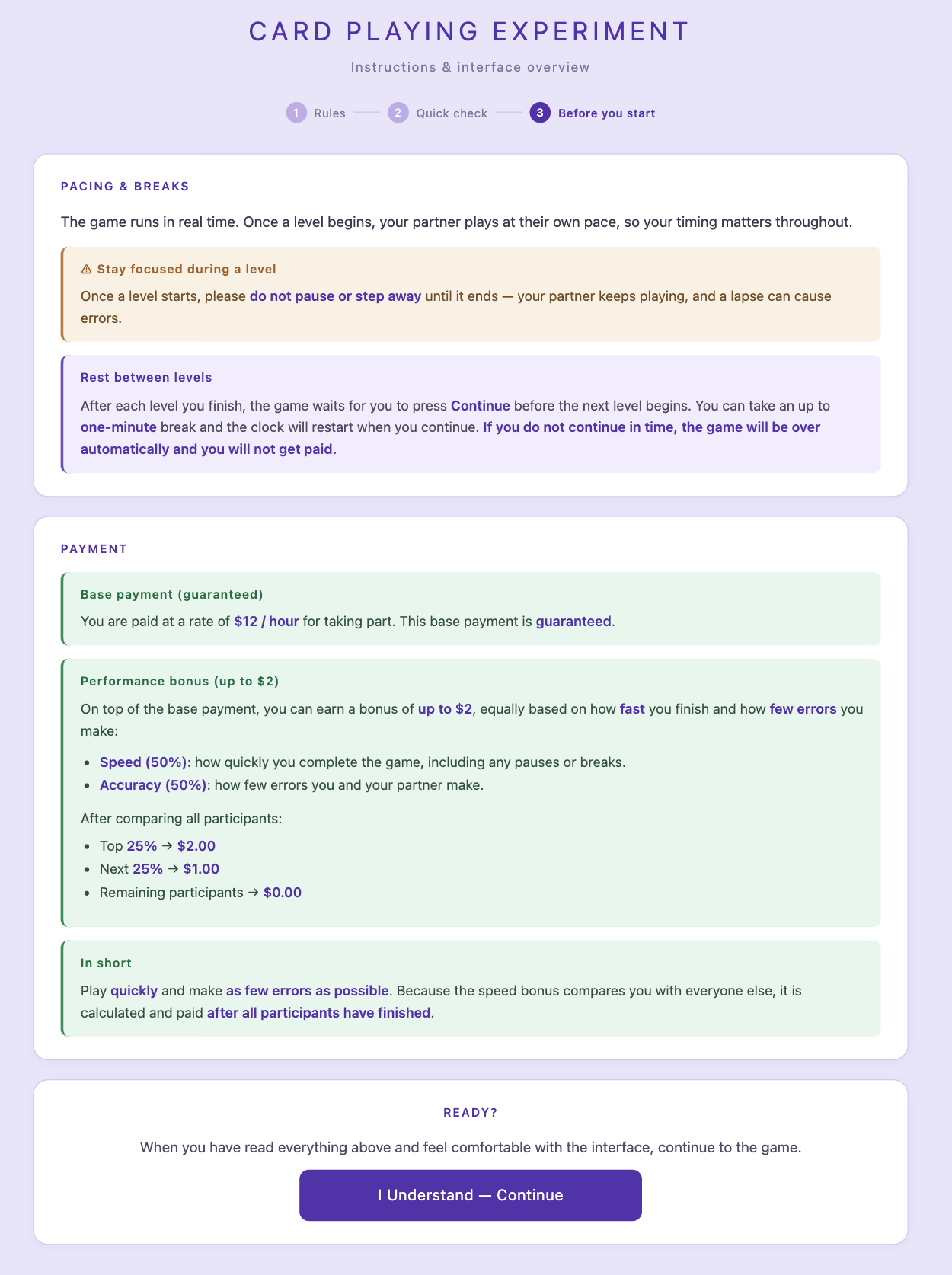}
    \caption{Instruction Page (3/3)}
    \label{fig:ins3}
\end{figure}

The game interface is shown in \Cref{fig:game_interface} and the instruction pages are shown in \Cref{fig:ins1,fig:ins2,fig:ins3}.

\section{LLM Experiments}
\subsection{Hyperparameters}
The hyperparameter settings are listed in \Cref{tab:hyperparameters}. Unless otherwise specified, we use the default settings.

\begin{table}[!th]
\tiny
\centering
\begin{adjustbox}{max width=\textwidth}
\begin{tabular}{p{0.30\textwidth}|p{0.62\textwidth}}
\toprule
Models & Settings \\
\midrule

\texttt{gemini-3.7-flash}
& thinking\_level: High; temperature: 1.0; TopP: 0.95; TopK: 64 \\

\texttt{gpt-5.6-luna}
& API model; default setting; thinking=medium \\

\texttt{gpt-6-astra}
& API model; default setting \\

\texttt{gpt-oss-120b}
& temperature: 0.7; max\_tokens: 10,000 \\

\texttt{gpt-oss-20b}
& temperature: 0.7; max\_tokens: 10,000 \\

\texttt{Qwen3-32B}
& temperature: 0.6; TopP: 0.95; TopK: 20; max\_tokens: 10,000 \\

\texttt{Qwen3-8B}
& temperature: 0.6; TopP: 0.95; TopK: 20; max\_tokens: 10,000 \\

\texttt{Llama-3.1-70B-Instruct}
& temperature: 0.6; max\_tokens: 10,000 \\

\texttt{Llama-3.1-8B-Instruct}
& temperature: 0.6; max\_tokens: 10,000 \\

\texttt{Llama-3.2-1B-Instruct}
& temperature: 0.6; max\_tokens: 10,000 \\

\bottomrule
\end{tabular}
\end{adjustbox}
\caption{Hyperparameter settings for the models.}
\label{tab:hyperparameters}
\end{table}

\newpage
\subsection{Prompt}
\label{app:prompt}
The game prompt is composed of two main parts: the game rule and the query. The query is composed of three parts: the most recent plays, the current game state, and the task instruction. Templates for these three parts across conditions are listed in \Cref{tab:prompt-conditions}.

\begin{tcolorbox}[
    colback=gray!8,
    colframe=black!60!blue,
    coltitle=white,
    fonttitle=\bfseries,
    title=Prompt of game rule,
    boxrule=1pt,
    arc=2pt,
    left=6pt, right=6pt, top=6pt, bottom=6pt,
    fontupper=\small  
]
You and your partner are playing a cooperative card game. This game is similar to a game called The Mind. However, our rules are slightly different, so please read the instructions very carefully.\\

            You and your partner play together as one team. 
            At each round (called a level), you and your partner are each dealt a new hand from a fresh deck of cards numbered 1 to 100. Cards from previous levels do not carry over or affect the cards dealt in the new level. You receive one card each in level 1, two cards each in level 2, and so on.\\

            Your shared goal is to arrange your and your partner's cards in ascending order, from lowest to highest. You cannot see your partner’s cards. You only know how many cards they have left and what cards have been played.\\

            How it works: 
            1. Each level deals you a new hand. In level N, you and your partner each hold N cards.
            2. You can only play your lowest card at any time. When you believe it is lower than every card your partner still holds, play it. There are no turns — whenever you sense your lowest card should go next, you play it.
            3. If you play a card while your partner still holds a lower one, that is an error — you played too early. When this happens the game pauses automatically, and every card lower than the one you just played is revealed and removed from your partner's hand. 
            4. Clear all cards to finish the level and move on. There are 10 levels in total.\\
            
            The performance of the game is equally based on **how fast you finish** and **how few errors you make**. 
            In other words, you should play QUICKLY and make as FEW errors as possible.

\end{tcolorbox}

\begin{tcolorbox}[
    colback=gray!8,
    colframe=black!60!blue,
    coltitle=white,
    fonttitle=\bfseries,
    title=Prompt of the query,
    boxrule=1pt,
    arc=2pt,
    left=6pt, right=6pt, top=6pt, bottom=6pt,
    fontupper=\small  
]
Below, you will see the 5 most recent plays in the game and the current game state. If the 5 most recent plays include the beginning of a new level, it will be clearly marked. Your task will be to decide whether to continue waiting or play your lowest card.\\

===5 MOST RECENT PLAYS===\\
-----BEGIN LEVEL X, DEAL NEW HANDS AND RESET CLOCK-----
\textit{[This exact line appears only when a new level begins.]}\\
\textit{[recent play history; format varies by prompt condition]}\\
===CURRENT GAME STATE===\\
\textit{[current hand, cards played, level info; format varies by prompt condition]}\\
===YOUR TASK===\\
\textit{[decision prompt, phrased per condition]}\\
\end{tcolorbox}

\begin{table}[!th]
\centering
\tiny
\begin{adjustbox}{max width=\textwidth}
\begin{tabular}{@{}p{1cm} p{3.5cm} p{5.5cm} p{3.8cm}@{}}
\toprule
\textbf{Condition} & \textbf{5 Most Recent Plays (Example)} & \textbf{Current Game State} & \textbf{Your Task} \\
\midrule
Time-based (1 second) &
Level: \texttt{<level>} \newline
Card played: \texttt{<card\_value>} \newline
Player: \texttt{<you\_or\_your\_partner>} \newline
Time since the start of level \texttt{<level>}: \texttt{<seconds>} second(s). &
You are currently on level \texttt{<level>}. \newline
Current hand: [\texttt{<card\_values>}] \newline
Your partner has \texttt{<num\_cards\_left>} card(s) left. \newline
The highest card played so far: \texttt{<highest\_card\_value>} \newline
You have been waiting for \texttt{<seconds>} second(s) since the last successfully played card \texttt{<card\_value>}. &
Based on the above information, do you want to wait \textcolor{blue}{one more second} or play your lowest card, i.e., \texttt{<card\_value>}?

Respond with exactly one word: \texttt{PLAY} or \texttt{WAIT}. \\

\midrule 

Time-based (0.5 second) &
Level: \texttt{<level>} \newline
Card played: \texttt{<card\_value>} \newline
Player: \texttt{<you\_or\_your\_partner>} \newline
Time since the start of level \texttt{<level>}: \texttt{<seconds>} second(s). &
You are currently on level \texttt{<level>}. \newline
Current hand: [\texttt{<card\_values>}] \newline
Your partner has \texttt{<num\_cards\_left>} card(s) left. \newline
The highest card played so far: \texttt{<highest\_card\_value>} \newline
You have been waiting for \texttt{<seconds>} second(s) since the last successfully played card \texttt{<card\_value>}. &
Based on the above information, do you want to wait \textcolor{blue}{another 0.5 second} or play your lowest card, i.e., \texttt{<card\_value>}?

Respond with exactly one word: \texttt{PLAY} or \texttt{WAIT}. \\

\midrule
Time-based (0.25 second) &
Level: \texttt{<level>} \newline
Card played: \texttt{<card\_value>} \newline
Player: \texttt{<you\_or\_your\_partner>} \newline
Time since the start of level \texttt{<level>}: \texttt{<seconds>} second(s). &
You are currently on level \texttt{<level>}. \newline
Current hand: [\texttt{<card\_values>}] \newline
Your partner has \texttt{<num\_cards\_left>} card(s) left. \newline
The highest card played so far: \texttt{<highest\_card\_value>} \newline
You have been waiting for \texttt{<seconds>} second(s) since the last successfully played card \texttt{<card\_value>}. &
Based on the above information, do you want to wait \textcolor{blue}{another 0.25 second} or play your lowest card, i.e., \texttt{<card\_value>}?

Respond with exactly one word: \texttt{PLAY} or \texttt{WAIT}. \\

\midrule

Time Estimate &
Level: \texttt{<level>} \newline
Card played: \texttt{<card\_value>} \newline
Player: \texttt{<you\_or\_your\_partner>} \newline
Time since the start of level \texttt{<level>}: \texttt{<seconds>} second(s). &
You are currently on level \texttt{<level>}. \newline
Current hand: [\texttt{<card\_values>}] \newline
Your partner has \texttt{<num\_cards\_left>} card(s) left. \newline
The highest card played so far: \texttt{<highest\_card\_value>} \newline
You have been waiting for \texttt{<seconds>} second(s) since the last successfully played card \texttt{<card\_value>}. &
Based on the information above, \textcolor{blue}{how many seconds would you like to wait before playing your lowest card, i.e., \texttt{<card\_value>}?
Respond with only a single numeric value. Do not include any explanation or units.} \\

\midrule 
\midrule 
Wait-based (count) & Level: \texttt{<level>} \newline
Card played: \texttt{<card\_value>} \newline
Player: \texttt{<you\_or\_your\_partner>} \newline
Decisions since the last played card: \texttt{<decision\_history>} \textit{\texttt{<decision\_history>} is a comma-separated sequence of each player's past decisions (e.g., \texttt{WAIT, WAIT, WAIT, WAIT, WAIT, PLAY})}
&
You are currently on level \texttt{<level>}. \newline
Current hand: [\texttt{<card\_values>}] \newline
Your partner has \texttt{<num\_cards\_left>} card(s) left. \newline
The highest card played so far: \texttt{<highest\_card\_value>} \newline
Your decision(s) since the last successfully played card \texttt{<card\_value>} are:\newline \texttt{<decision\_history>}. 
&
Based on the above information, \textcolor{blue}{do you want to wait one more time or play your lowest card, i.e., \texttt{<card\_value>}?} Respond with exactly one word: \texttt{PLAY} or \texttt{WAIT}\\
\midrule 
Wait-based (count) & Level: \texttt{<level>} \newline
Card played: \texttt{<card\_value>} \newline
Player: \texttt{<you\_or\_your\_partner>} \newline
Number of WAITs since the last played card \texttt{<card\_value>}: \texttt{<num\_of\_waits>}.
&
You are currently on level \texttt{<level>}. \newline
Current hand: [\texttt{<card\_values>}] \newline
Your partner has \texttt{<num\_cards\_left>} card(s) left. \newline
The highest card played so far: \texttt{<highest\_card\_value>} \newline
You have chosen to WAIT \texttt{<num\_of\_waits>} times since the last successfully played card \texttt{<card\_value>}.
&
Based on the above information, \textcolor{blue}{do you want to wait one more time or play your lowest card, i.e., \texttt{<card\_value>}?} Respond with exactly one word: \texttt{PLAY} or \texttt{WAIT}\\

\bottomrule
\end{tabular}
\end{adjustbox}
\caption{Prompt template shared across conditions, with condition-specific content for each templated section.}
\label{tab:prompt-conditions}
\end{table}

\newpage
\section{Complementary Results}
\change{\subsection{Self-report analysis of human participants}
\label{app:self_report}
After the game, participants were asked to describe the strategy they used (open-ended question: \textit{Describe your strategy, or write NA}). We coded the responses into the categories shown in Table~\ref{tab:self_report}. Most participants described their strategy vaguely or in terms of intuition, and few described explicit counting or reasoning about their partner's hand. This is consistent with the limits of introspective access to decision processes \citep{nisbett1977telling}, and motivates our reliance on computational modeling of behavior. 

\begin{table}[h]
\centering
\small
\begin{tabular}{p{4.2cm}rp{7cm}}
\toprule
\textbf{Strategy category} & \textbf{N}  & \textbf{Example response} \\
\midrule
Intuition/no explicit strategy & 13&  \textit{Just try to play my cards when it felt right.} \\
Wait longer for higher cards &35 &\textit{I tried to click as soon as I saw a number closest to my lowest number} \\
Attend to partner's timing & 33&  \textit{I just waited for my partner to put a card down. If the partner did not in a reasonable time, I figured my card was probably lower} \\
Counting / explicit timing &10 &  \textit{I would count 5 seconds before placing a card when waiting on my partner if we were in a sort of a deadlock}\\
Other / uninformative & 23& \textit{The other guy was pretty fast, so I didn't really have time to use strategy} \\
NA & 36 & \textit{NA} \\
\midrule
Total & 150 & \\
\bottomrule
\end{tabular}
\caption{Summary of participants' self-reported strategies. Responses were coded into categories by one of the authors. Some responses could fall into multiple categories.}
\label{tab:self_report}
\end{table}
We also observe that some participants reported not only adapting to to their partner, but also attempting to influence their partner's behavior (E.g., \textit{Tried to get my partner to slow down by pausing between cards.})

}

\subsection{BIC}
\label{bic}
\begin{figure}[!th]
    \centering
    \includegraphics[width=\linewidth]
    %{figs/final_alltogether_bic_condensed.png}
    {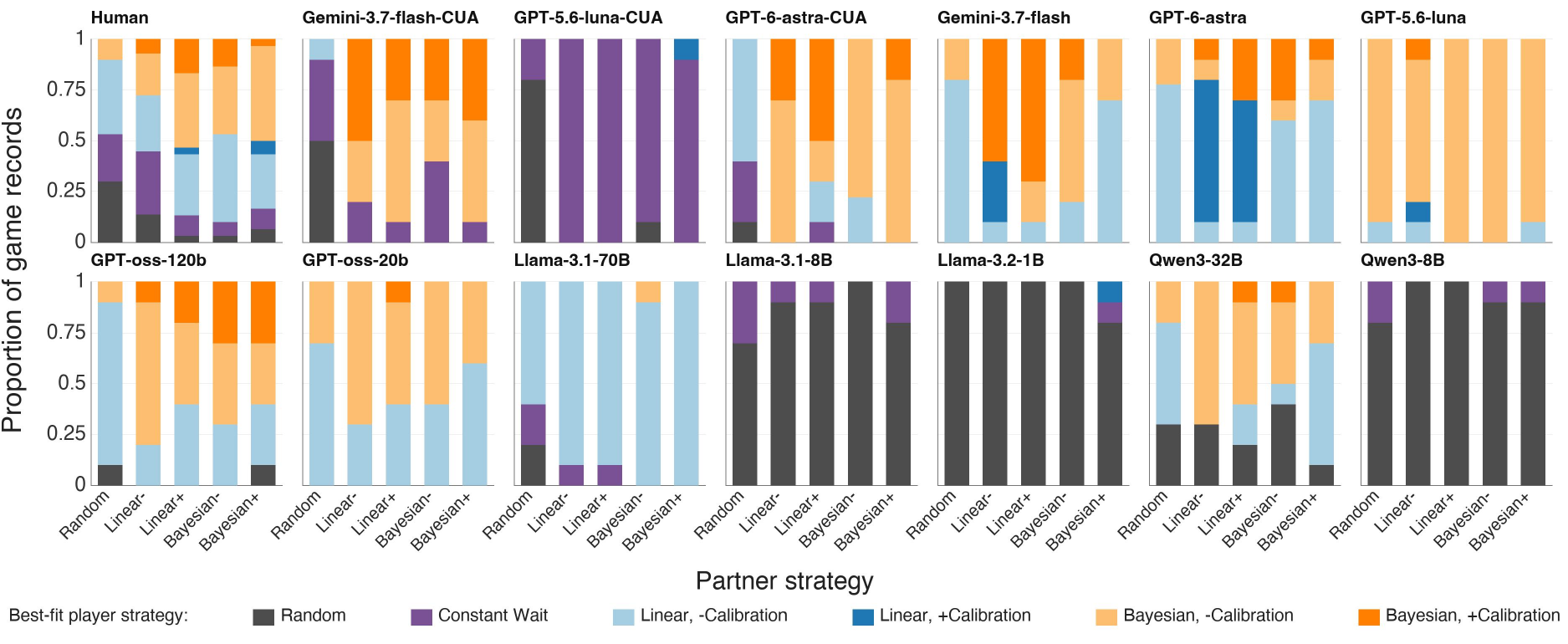}
    \caption{Proportion of Human, LLM, and CUA Game Records Best Fit by Strategy Models (BIC)}
    \label{alltogether_bic}
\end{figure}

As shown in \Cref{alltogether_bic}, where the AIC and BIC analysis differed, BIC preferentially selected simpler Random or Linear without calibration models. 

\subsection{Efficiency comparison across players}
\begin{figure}[!th]
    \centering
    \includegraphics[width=\linewidth]{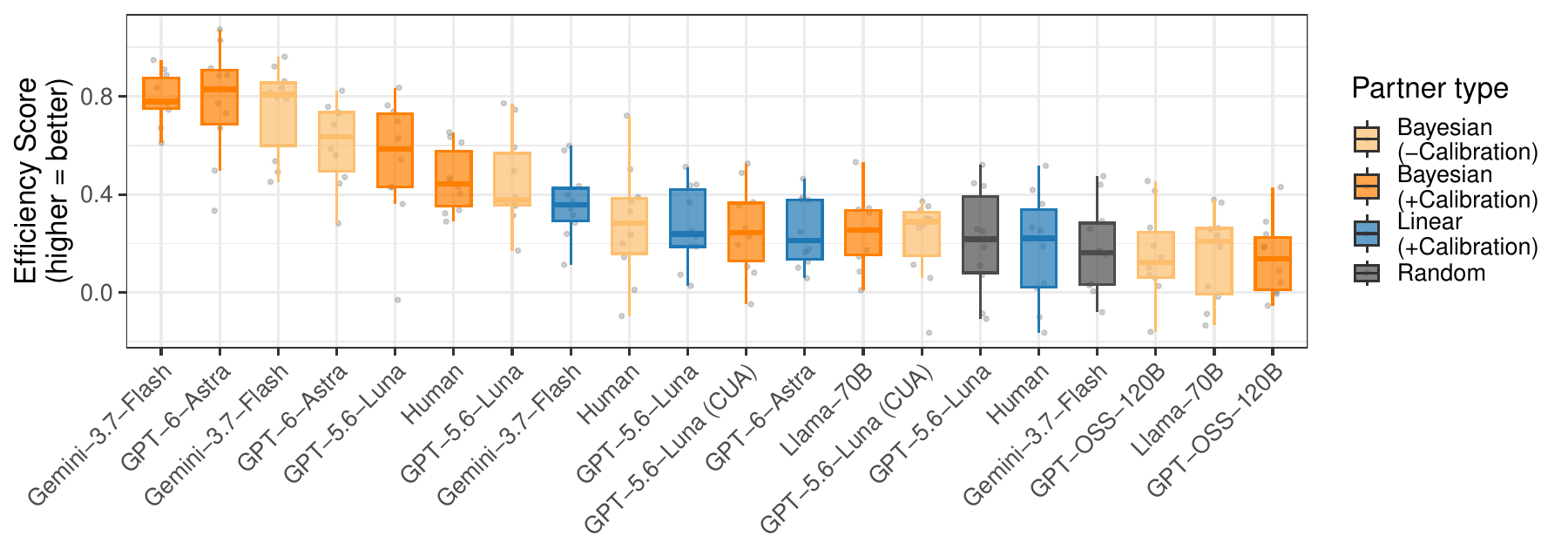}
    \caption{Top 20 most efficient player-partner combinations.} % in Experiment 1}
    \label{fig:top20}
\end{figure}

The top-20 most efficient player-partner combination is reported in \Cref{fig:top20}.

\subsection{Time consistency and CV of time of models across prompt types}
\label{app:time}
\begin{figure}[!th]
        \centering
        \includegraphics[width=0.6\linewidth]{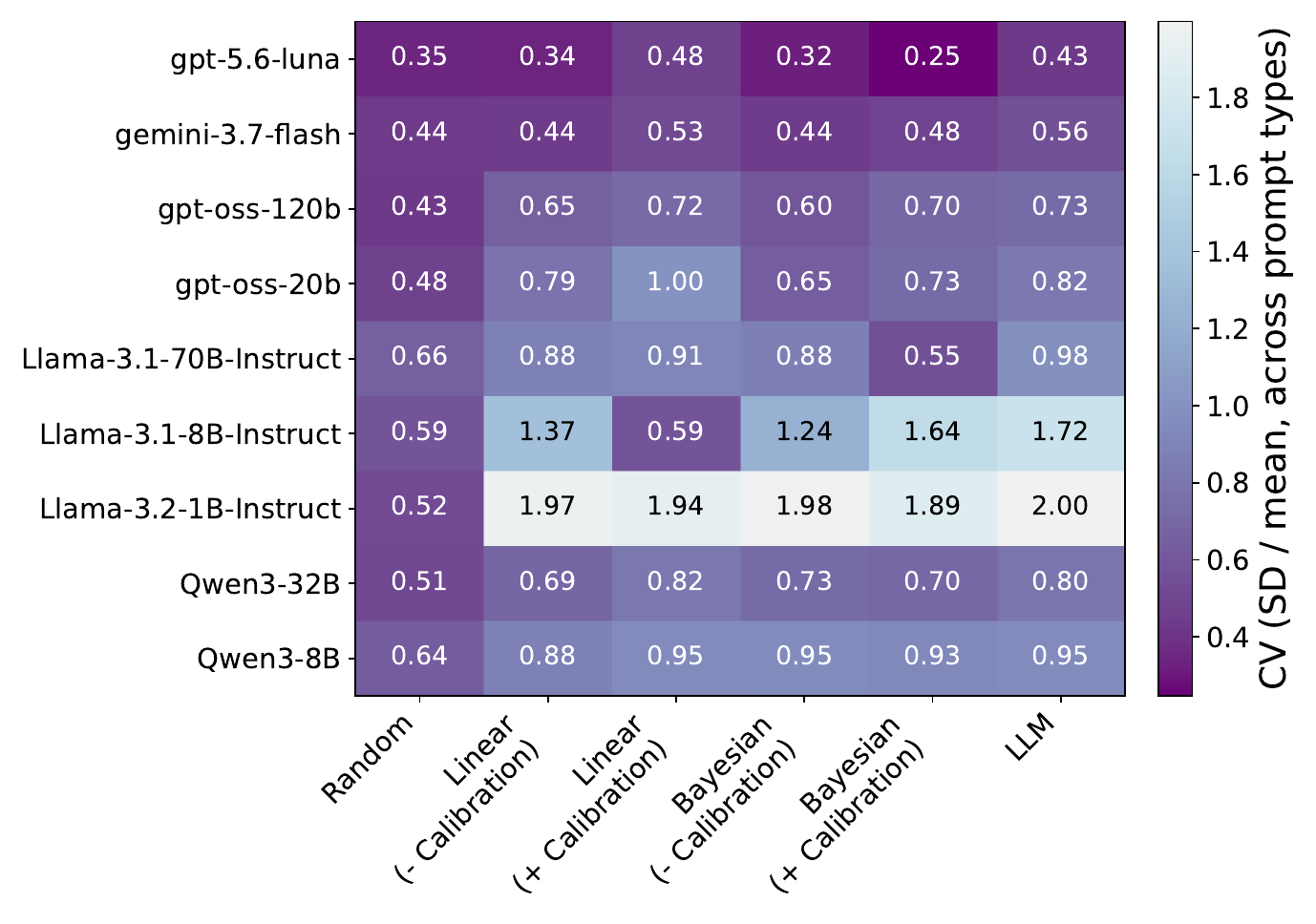}
        \caption{Coefficient of variation across prompt types.}
        \label{fig:cv}
    \end{figure}

\Cref{fig:cv} shows the Coefficient of variation of different models across prompt types.

\begin{figure}
    \centering
    \includegraphics[width=\linewidth]{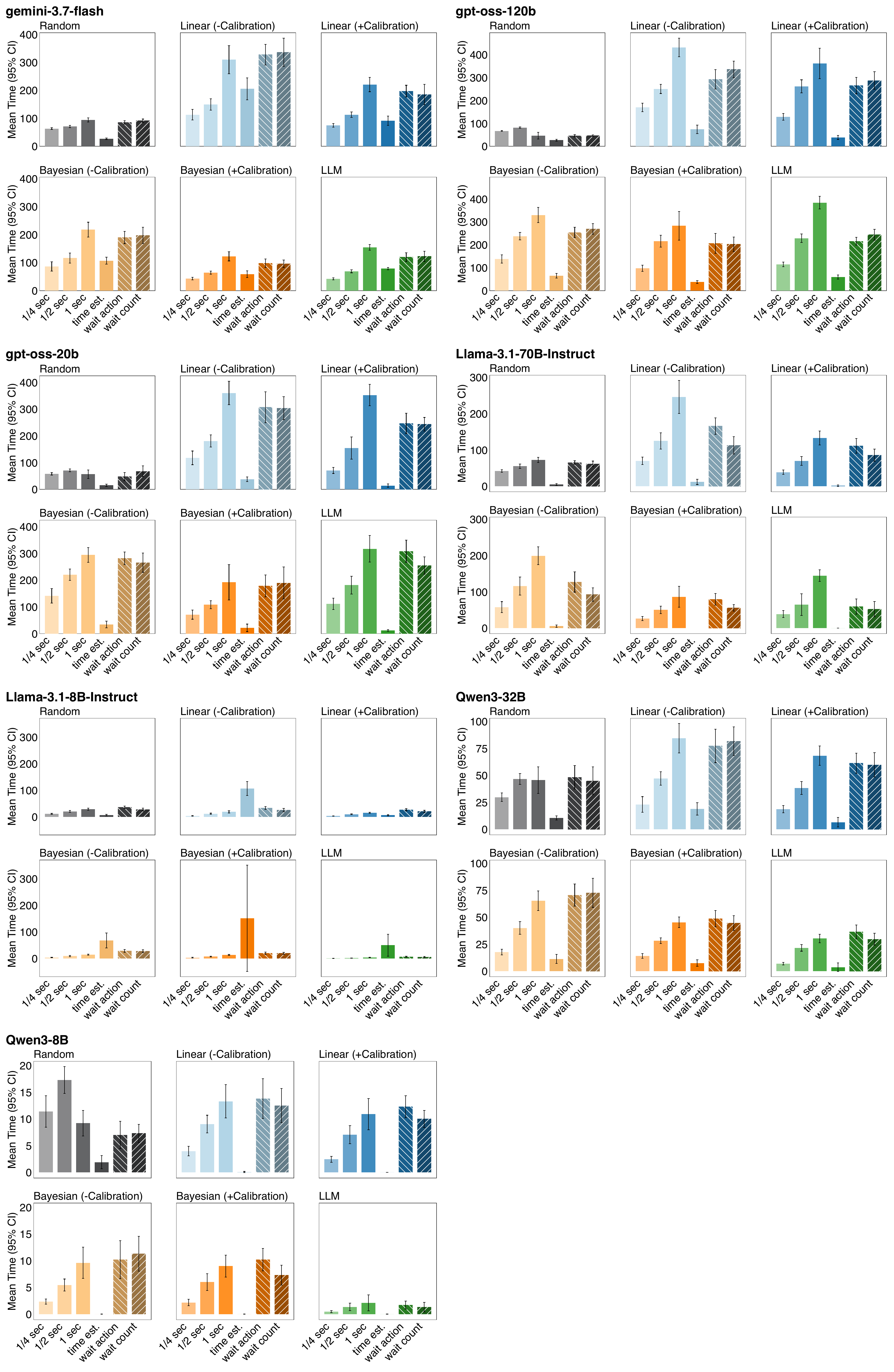}
    \caption{Mean play time by partner type and prompt type for the LLMs. Error bars show 95\% CI.}
    \label{fig:time_other}
\end{figure}

\Cref{fig:time_other} shows the time consistency of other models that are not reported in the main text. We find similar patterns: simply changing the temporal granularity used to represent waiting time
in the prompt can lead to substantially different behavioral patterns.

\subsection{Statistical Analysis of Player's Adaptive Behavior}
\begin{figure}[!th]
    \centering
    \begin{subfigure}{0.48\textwidth}
        \centering
        \includegraphics[width=\textwidth]{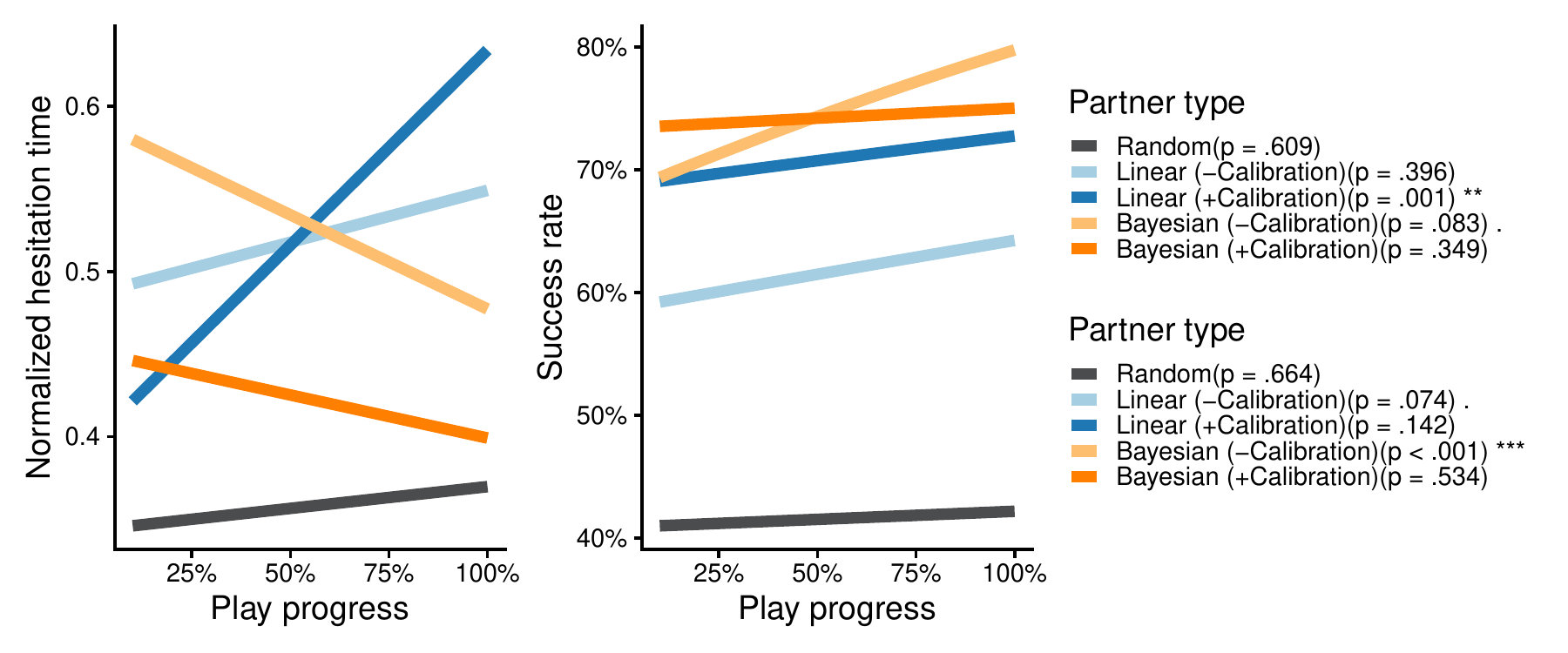}
        \caption{Human}
        \label{fig:human-stats}
    \end{subfigure}
    \hfill
    \begin{subfigure}{0.48\textwidth}
        \centering
        \includegraphics[width=\textwidth]{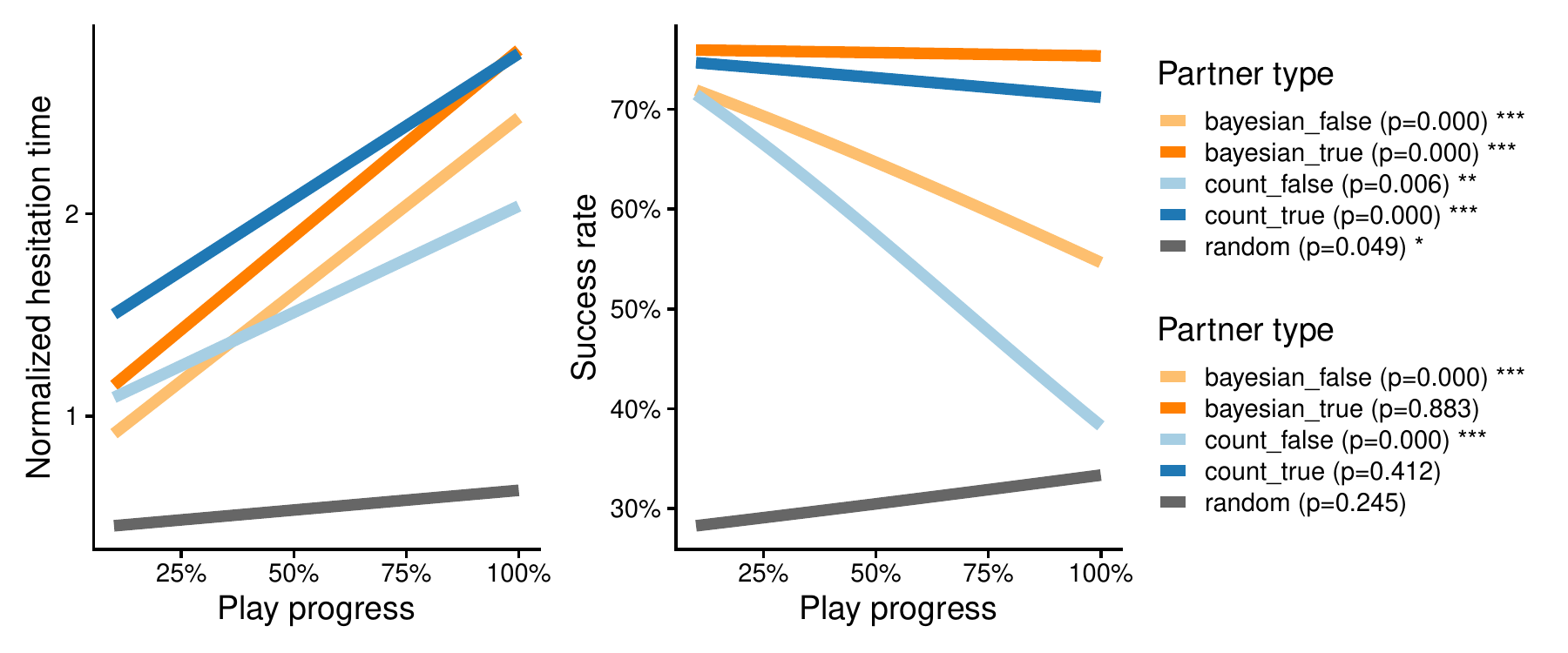}
        \caption{Gemini-3.7-Flash (CUA)}
        \label{fig:gemini-stats}
    \end{subfigure}
    
    \begin{subfigure}{0.48\textwidth}
        \centering
        \includegraphics[width=\textwidth]{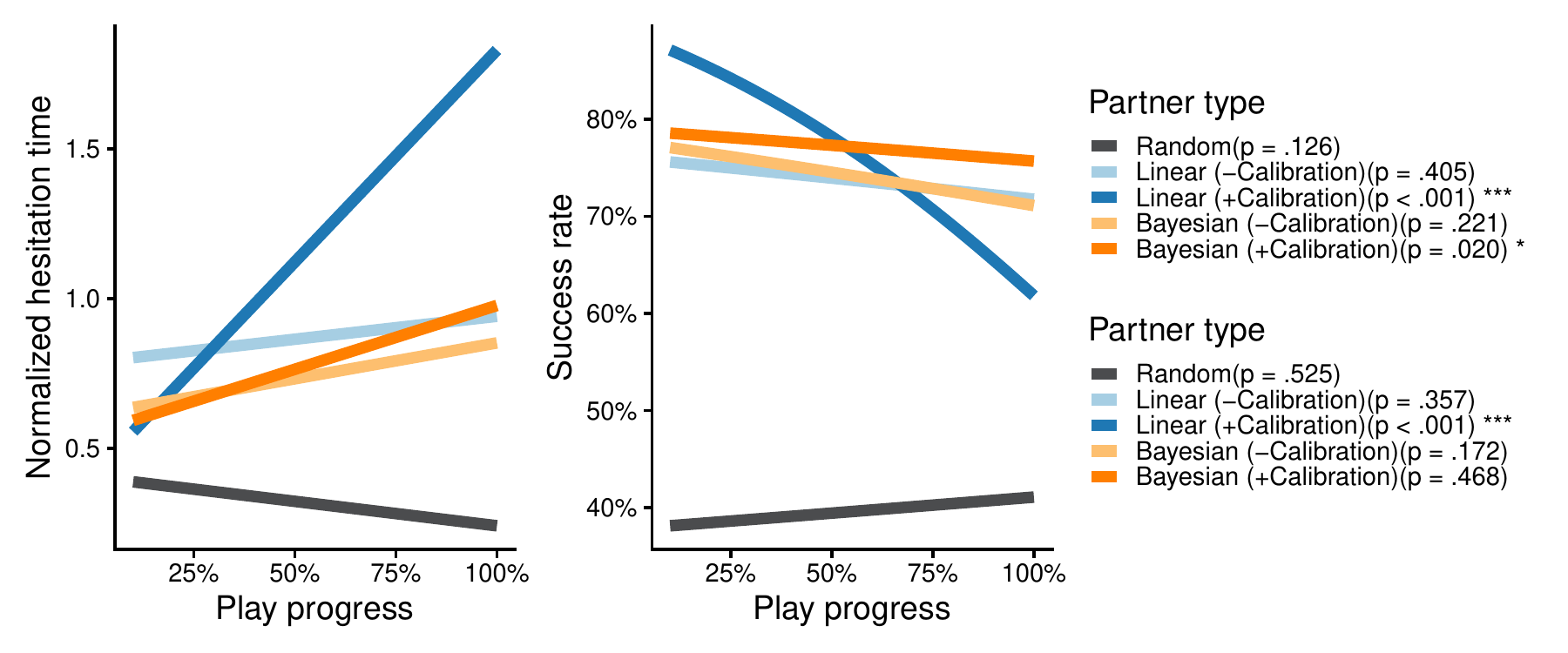}
        \caption{GPT-6-astra (CUA)}
        \label{fig:stats-gpt6}
    \end{subfigure}
    \hfill
\begin{subfigure}{0.48\textwidth}
        \centering
        \includegraphics[width=\textwidth]{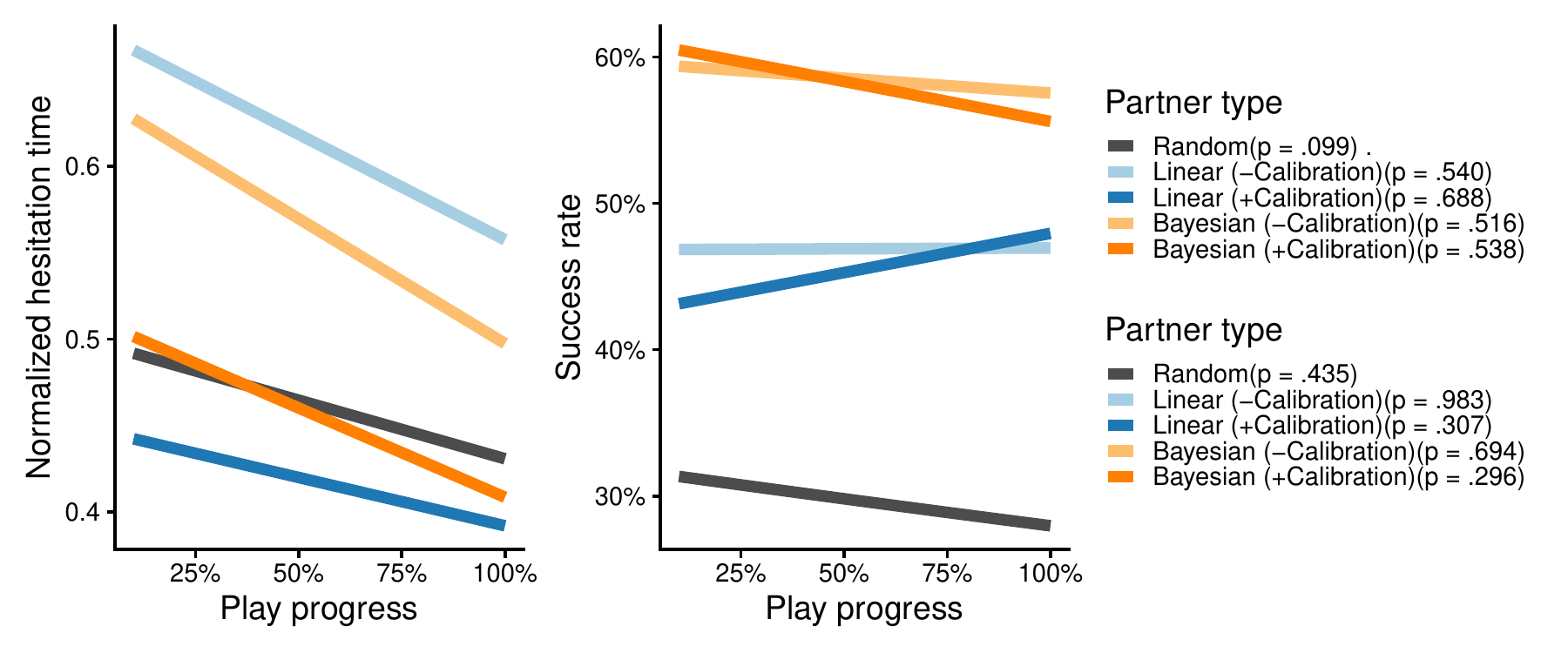}
        \caption{GPT-5.6-luna (CUA)}
        \label{fig:stats-gptluna}
    \end{subfigure}
    \caption{Fitted regression lines for normalized hesitation time (left) and success rate (right) over the course of a hand, for humans and three CUAs, by partner type.}
    \label{fig:stats-realtime}
\end{figure}

\begin{figure}
    \centering
    \includegraphics[width=\linewidth]{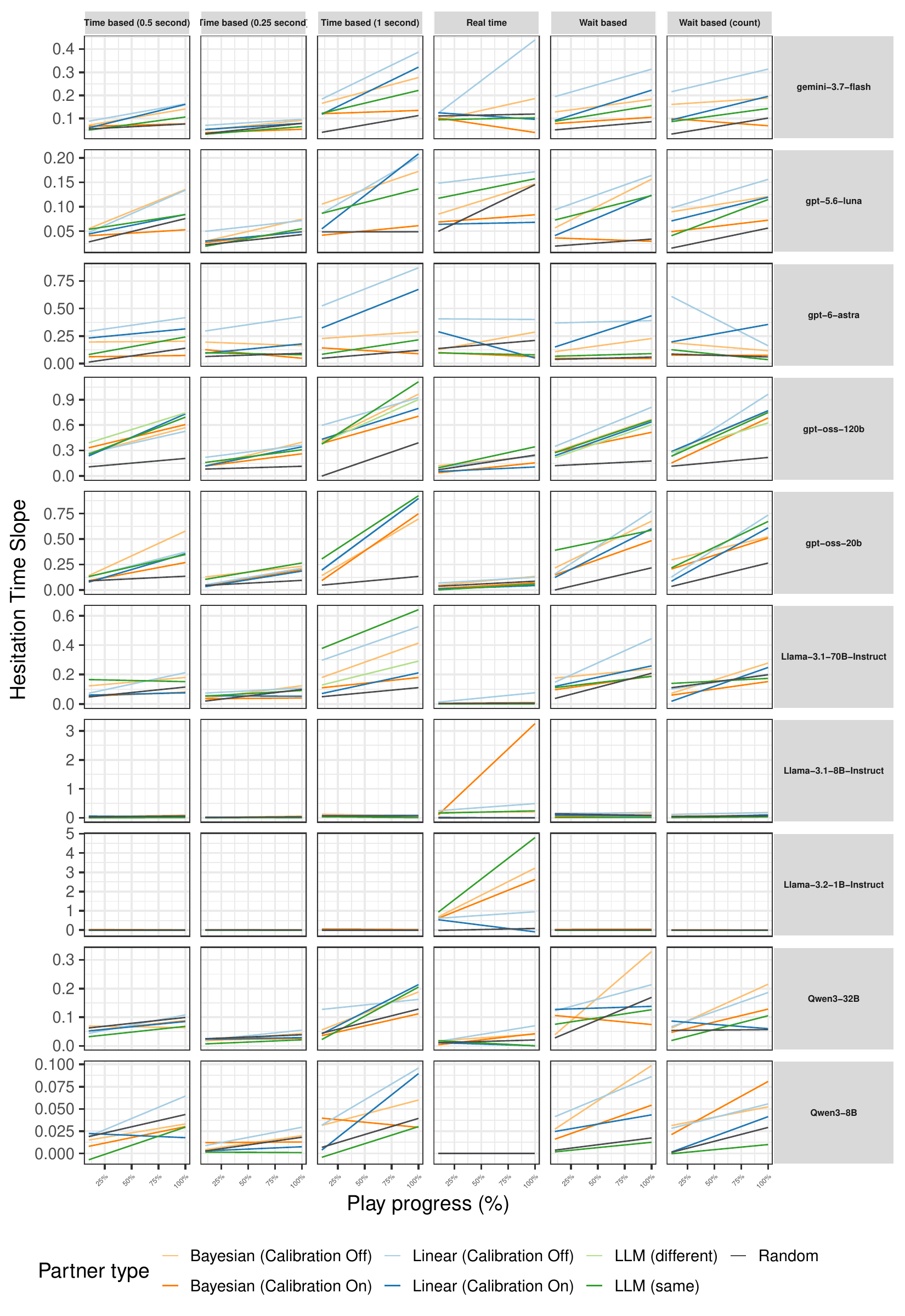}
    \caption{The hesitation slope from the mixed model for each model by prompt setting.}
    \label{fig:time_slope}
\end{figure}

\begin{figure}
    \centering
    \includegraphics[width=\linewidth]{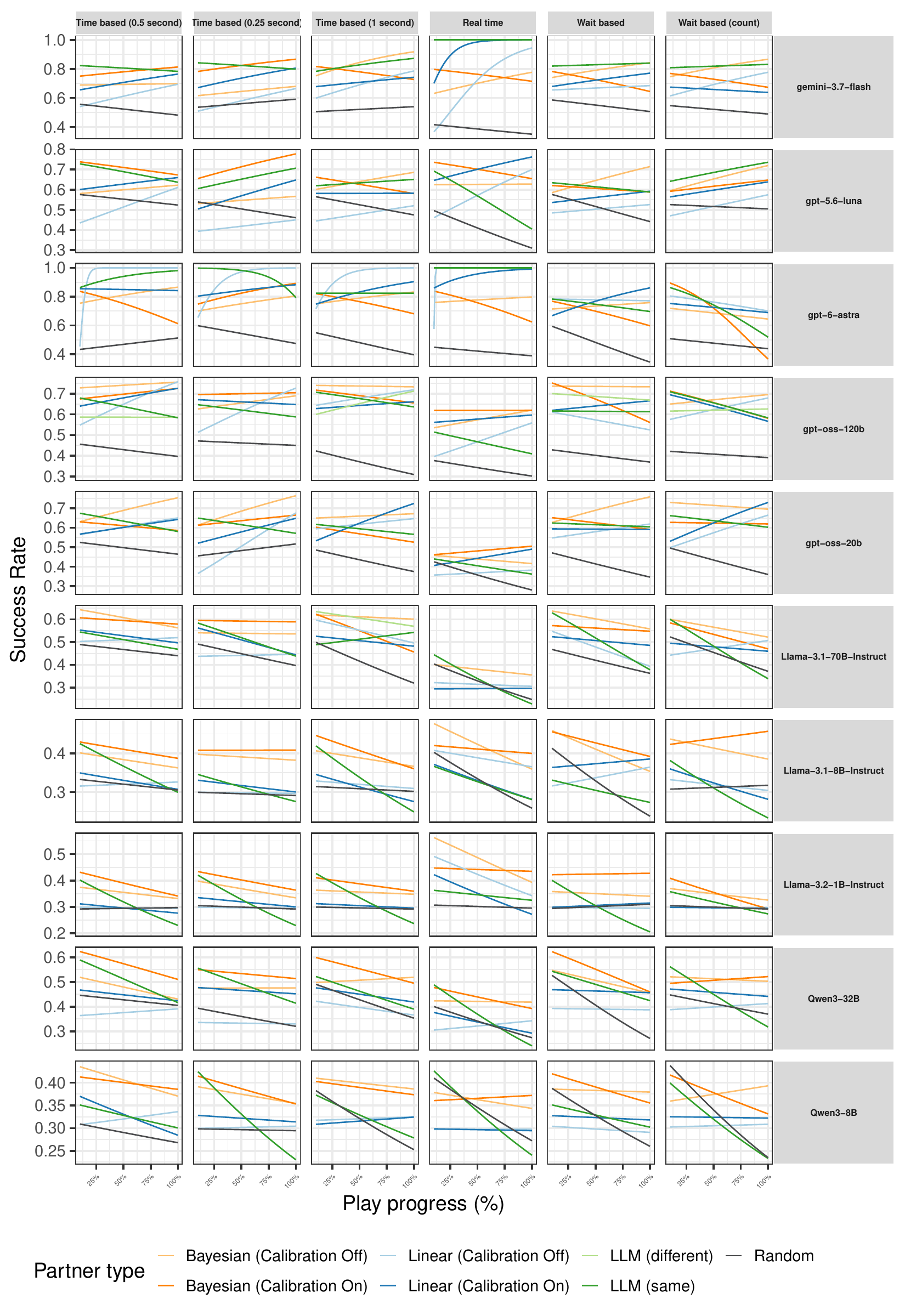}
    \caption{The success slope from the mixed model for each model by prompt setting.}
    \label{fig:sus}
\end{figure}

\Cref{fig:time_slope,fig:sus} show the fitted regression lines for normalized hesitation time and success rate for each LLM and prompt combination respectively. \Cref{fig:stats-realtime} visualizes the fitted regression lines for normalized hesitation time and success rate over the course of a hand, for humans and three CUAs. We find that (i) most prompted-based LLMs show slower response time and lower success rate as time goes; (ii) even in more comparable human-CUA settings, only humans playing with Bayesian Player (without Calibration) show both (marginally) statistically significant faster speed and higher success rate over the course of the game.

\label{app:stats}

\end{document}